%% file: authordraft.tex
\PassOptionsToPackage{table}{xcolor}
\documentclass[sigconf]{acmart}

\usepackage{verbatim}
\usepackage{algorithm}
\usepackage{algorithmic}
\usepackage{multirow}

\AtBeginDocument{%
  }

\setcopyright{acmlicensed}
\copyrightyear{2018}
\acmYear{2025}
\acmDOI{XXXXXXX.XXXXXXX}
\acmConference[MM '25]{33st ACM International Conference on Multimedia}{October 27--31,
  2025}{Dublin, Ireland}
\acmISBN{978-1-4503-XXXX-X/2018/06}

\acmSubmissionID{123-A56-BU3}

\begin{document}

\title{Graph-Guided Dual-Level Augmentation for 3D Scene Segmentation}

\author{Hongbin Lin}
\authornote{Both authors contributed equally to this research.}
\affiliation{%
  \institution{The Hong Kong University of Science and Technology (Guangzhou)}
  \city{Guangzhou}
  \country{China}
}
\email{hlin199@connect.hkust-gz.edu.cn}

\author{Yifan Jiang}
\authornotemark[1]
\affiliation{%
  \institution{The Hong Kong University of Science and Technology (Guangzhou)}
  \city{Guangzhou}
  \country{China}
}
\email{yjiang578@connect.hkust-gz.edu.cn}

\author{Juangui Xu}
\authornotemark[1]
\affiliation{%
  \institution{The Hong Kong University of Science and Technology (Guangzhou)}
  \city{Guangzhou}
  \country{China}
}
\email{juanguixu@163.com}

\author{Jesse J. Xu}
\affiliation{%
  \institution{University of Toronto}
  \city{Toronto}
  \country{Canada}
}
\email{jessejiaxi.xu@mail.utoronto.ca}

\author{Yi Lu}
\affiliation{%
  \institution{The Hong Kong University of Science and Technology (Guangzhou)}
  \city{Guangzhou}
  \country{China}
}
\email{yilo6117@gmail.com}

\author{Zhengyu Hu}
\affiliation{%
  \institution{The Hong Kong University of Science and Technology (Guangzhou)}
  \city{Guangzhou}
  \country{China}
}
\email{huzhengyu477@gmail.com}

\author{Ying-Cong Chen}
\authornote{Corresponding author.}
\affiliation{%
  \institution{The Hong Kong University of Science and Technology (Guangzhou)}
  \city{Guangzhou}
  \country{China}
}
\email{yingcongchen@hkust-gz.edu.cn}

\author{Hao Wang}
\authornotemark[2]
\affiliation{%
  \institution{The Hong Kong University of Science and Technology (Guangzhou)}
  \city{Guangzhou}
  \country{China}
}
\email{haowang@hkust-gz.edu.cn}

\renewcommand{\shortauthors}{H. Lin et al.}

\begin{abstract}
3D point cloud segmentation aims to assign semantic labels to individual points in a scene for fine-grained spatial understanding. Existing methods typically adopt data augmentation to alleviate the burden of large-scale annotation. However, most augmentation strategies only focus on local transformations or semantic recomposition, lacking the consideration of global structural dependencies within scenes. To address this limitation, we propose a graph-guided data augmentation framework with dual-level constraints for realistic 3D scene synthesis. Our method learns object relationship statistics from real-world data to construct guiding graphs for scene generation. Local-level constraints enforce geometric plausibility and semantic consistency between objects, while global-level constraints maintain the topological structure of the scene by aligning the generated layout with the guiding graph. Extensive experiments on indoor and outdoor datasets demonstrate that our framework generates diverse and high-quality augmented scenes, leading to consistent improvements in point cloud segmentation performance across various models. Code is available at: https://github.com/alexander7xu/DualLevelAug
\end{abstract}





\maketitle

\input{sec/1_intro}

\input{sec/2_relatedwork}

\input{sec/3_methodology}

\input{sec/4_experiment}
\input{sec/5_conclusion}
\input{sec/6_Acknowledgment}

\bibliographystyle{ACM-Reference-Format}
\bibliography{base}
\newpage

\clearpage

\appendix
\newpage
\input{sec/appendix}

\end{document}

%% file: sec/1_intro.tex
\section{Introduction}
\label{sec:intro}

\begin{figure}[t]
    \centering
    \includegraphics[width=0.95\linewidth]{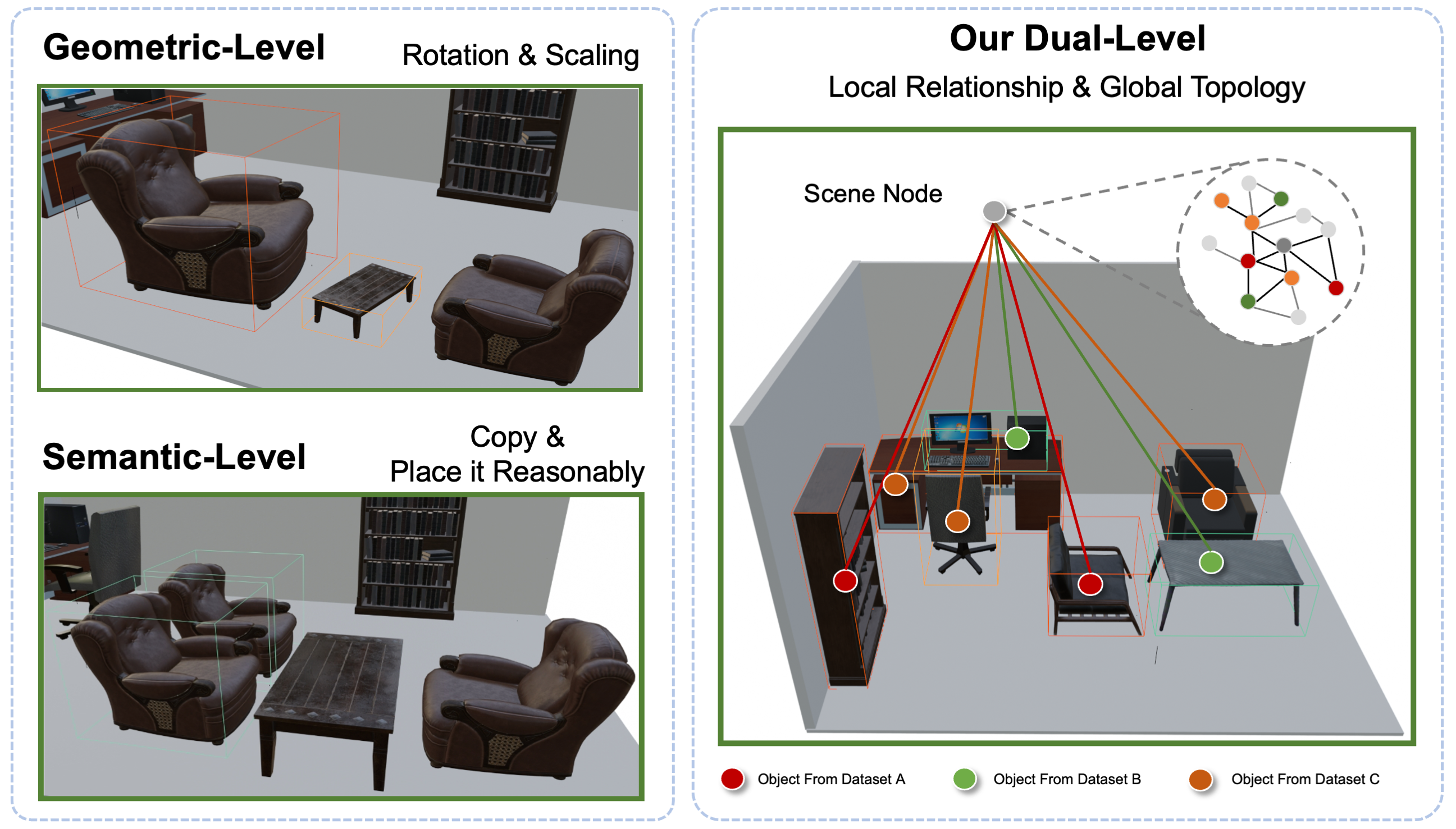}
    \vspace{-0.5cm}
    \caption{Single-Level Augmentation vs. Dual-Level Augmentation. Traditional approaches (left) typically operate at the local level. They either apply simple geometric transformations, such as rotation and scaling. Or they perform semantic-level manipulations by inserting copied objects into contextually reasonable locations. In contrast, our method (right) combines both local and global constraints, while also enabling cross-scene object integration. By modeling the scene as a graph, our method supports more complex and high-quality scene rearrangements, enhancing diversity and semantic coherence in synthesized point cloud data.}
    \label{fig:intro}
\vspace{-12pt}
\end{figure}

Point cloud segmentation, which aims to assign semantic or instance labels to each 3D point in a scene, is a fundamental task in 3D scene understanding. It is crucial in numerous applications such as robotic navigation, augmented reality, autonomous driving, and digital twin systems~\cite{qi2017pointnet,guo2020deep,ma2019multi}. Recent advancements leverage transformer architectures, graph-based reasoning, and multi-modal fusion to enhance segmentation performance further~\cite{khan2022transformers, zhao2021point,sarker2024comprehensive}. Despite these successes, existing methods rely heavily on large-scale annotated datasets, which are costly to acquire and label, posing a bottleneck for further progress~\cite{xiao2024survey,li2022improving,hu2024let,hu2023leveraging}.

Real-world 3D scenes, whether indoor environments or outdoor streetscapes, exhibit complex spatial arrangements of objects. These arrangements are not random but follow intricate underlying distributions governed by physical laws, functional requirements, semantic context, and common usage patterns—such as vehicles on roads or furniture in rooms~\cite{armeni20163d, dai2017scannet, hu2020randla, wijayathunga2023challenges}. Learning these distributions is essential for robust scene understanding, especially in tasks like point cloud segmentation. However, due to the high cost of acquiring and annotating large-scale 3D datasets, models often suffer from limited exposure to diverse spatial configurations.

Data augmentation is widely adopted to mitigate this limitation, enrich training data, and improve generalization. However, a key challenge remains: \textit{ How to ensure that the augmented samples align with the structural and semantic constraints observed in real-world environments?} Without such constraints, synthesized data may introduce unrealistic object arrangements or implausible spatial relationships, which can mislead the model and hinder its performance when deployed in real scenarios~\cite{wang2024comprehensive,zhu2024advancements}.

Existing point cloud data augmentation techniques can be broadly classified into two categories: geometric-level and semantic-level methods. Geometric-level methods, such as rotation, scaling, and jittering~\cite{chen2020pointmixup, li2020pointaugment, nekrasov2021mix3d}, primarily introduce local perturbations. While useful, they typically fail to generate novel scene layouts and thus explore only a limited region of the underlying scene distribution. Semantic-level methods—including generative models~\cite{achlioptas2018learning, yang2019pointflow, lehner20223d}and object insertion or replacement strategies~\cite{gong2022neighborhood,ren2022object}—attempt to modify scene composition more globally. However, they face difficulties in maintaining semantic consistency and physical realism, producing configurations that lie outside the target distribution of valid scenes. A key limitation persists across these methods. They lack explicit mechanisms to handle the complex distributions inherent in real-world 3D data. Consequently, they fail to accurately model or rigorously enforce these crucial relational and geometric regularities. Moreover, both geometric-level and semantic-level methods predominantly focus on local constraints, without considering the global structural dependencies or topological relationships that are critical for realistic and coherent scene generation~\cite{kolbe2021semantic}.

To overcome these limitations, we propose a novel data augmentation framework that synthesizes realistic and diverse 3D point cloud scenes by enforcing dual-level constraints. Our method explicitly models object co-occurrence statistics and spatial relationships from real-world datasets to guide scene generation with both local and global structural coherence. We first construct abstract guiding graphs that encode the desired scene topology, where the node activation is regulated by Jensen-Shannon (JS) divergence\cite{lin1991divergence} to ensure consistency with the object category distribution in the training data. Objects are then placed into the scene and refined through a constraint-driven optimization process. Local-level constraints enforce geometric plausibility and semantic consistency, such as collision avoidance and functional relationships, while global-level constraints preserve the overall scene structure by minimizing the Graph Global Constraint Loss (GGCL) between the generated graph and the guiding graph. This dual-level strategy enables the generation of high-quality augmented scenes that capture complex spatial dependencies, effectively enhancing model robustness for point cloud segmentation.

The contributions of this work are summarized as follows:

\textbf{1) A graph-guided synthesis framework} that models object co-occurrence statistics and spatial relationships to generate diverse and semantically meaningful 3D scenes.
    
\textbf{2) A dual-level constraint optimization strategy} that enforces geometric and semantic consistency at the local level, while maintaining global topological regularity via graph structure alignment.
    
\textbf{3) Extensive experimental validation} demonstrating that our method significantly improves segmentation performance across indoor and outdoor datasets, outperforming conventional augmentation techniques.

%% file: sec/2_relatedwork.tex
\section{Related work}
\label{sec:relatedwork}

\textbf{Point Cloud Segmentation.}
Point cloud segmentation, which aims to assign semantic or instance labels to 3D points, is a core task in 3D scene understanding with applications in robotics, autonomous driving, and digital twin systems~\cite{qi2017pointnet,guo2020deep}. It can be broadly categorized into indoor and outdoor segmentation. Indoor segmentation methods typically operate on structured yet cluttered environments such as offices, homes, and classrooms, where object categories are diverse and spatial arrangements are irregular. Point-based models~\cite{qi2017pointnet++, qian2022pointnext, thomas2019kpconv, lei2020spherical} extract local geometric features directly from raw point clouds. More recently, transformer-based models~\cite{lai2022stratified, zhou2020cylinder3d, zhao2021point, wu2022point, wang2023octformer, yang2025swin3d, wu2025sonata} have achieved state-of-the-art performance by capturing long-range dependencies and integrating hierarchical spatial context. In contrast, outdoor segmentation methods target large-scale scenes like streets and highways, which exhibit more regular geometric patterns and stronger layout priors. Range-view-based approaches~\cite{wu2018squeezeseg, wu2019squeezesegv2, cortinhal2020salsanext} project LiDAR data into 2D for efficient processing. Meanwhile, voxel-based and hybrid representations~\cite{tang2020searching, choy20194d, peng2024oa} leverage sparse convolutions or aggregation-based mechanisms to handle scalability and preserve fine-grained structure. Despite progress in both domains, limited training diversity and strong dataset biases remain major challenges, motivating the need for structured data augmentation strategies.

\textbf{Augmentation for Point Clouds.}
Data augmentation has been widely explored across point cloud tasks, including classification~\cite{li2020pointaugment}, detection~\cite{chen2023voxelnext}, and registration~\cite{zheng2024regiformer}. For segmentation, augmentation is particularly important due to the high cost of annotating dense 3D scenes. Classic techniques apply geometric perturbations such as rotation, jittering, and scaling~\cite{qi2017pointnet, chen2020pointmixup, kim2021point, lai2022stratified, zhou2020cylinder3d, zhang2022pointcutmix, xiao2022polarmix}. Recent advances propose learning-based augmentation strategies~\cite{li2020pointaugment, nekrasov2021mix3d} optimize augmentations through policy search or region-level mixing. While effective to some extent, most of these techniques focus on local geometry and are agnostic to the spatial relationships or functional roles of objects. Moreover, they often ignore the topological or contextual semantics that govern real-world 3D layouts, limiting their effectiveness in complex indoor or multi-object scenes. This motivates structured augmentation strategies that model relationships beyond individual objects.

\textbf{Semantic-Aware Scene Composition.}
Moving beyond isolated object augmentation, semantic-aware scene composition aims to synthesize or manipulate entire 3D scenes while preserving realistic spatial arrangements. Early work employed rule-based layout priors or scene grammars~\cite{fisher2012example, liu2018semi, lim2022point, xiang2024synthetic}, while more recent approaches leverage generative models such as GANs~\cite{achlioptas2018learning}, VAEs~\cite{yang2019pointflow}, and diffusion models~\cite{chen2023diffindscene,zhai2023commonscenes,tang2024diffuscene,gao2024graphdreamer,wang2024echoscene,jiang2024lexical} to synthesize entire indoor scenes. These methods often incorporate scene graphs~\cite{zhou2019scenegraphnet, armeni20163d} to encode object co-occurrence and spatial relations. However, generative approaches still face challenges in aligning with real-world distributions, especially in segmentation-specific settings where fine-grained point-level geometry and contextual structure matter. In parallel, object-level composition methods~\cite{gong2022neighborhood, ren2022object} propose inserting or rearranging objects based on proximity or class affinity, but often oversimplify the semantics of spatial configurations. Our work propose a dual-level constraint framework that combines local physical and semantic relationships with global topological guidance, enabling the generation of diverse, plausible scenes tailored for point cloud segmentation.

%% file: sec/3_methodology.tex
\section{Methodology}
\label{sec:methodology}

\begin{figure*}[htbp]
    \centering
    \includegraphics[width=0.90\textwidth]{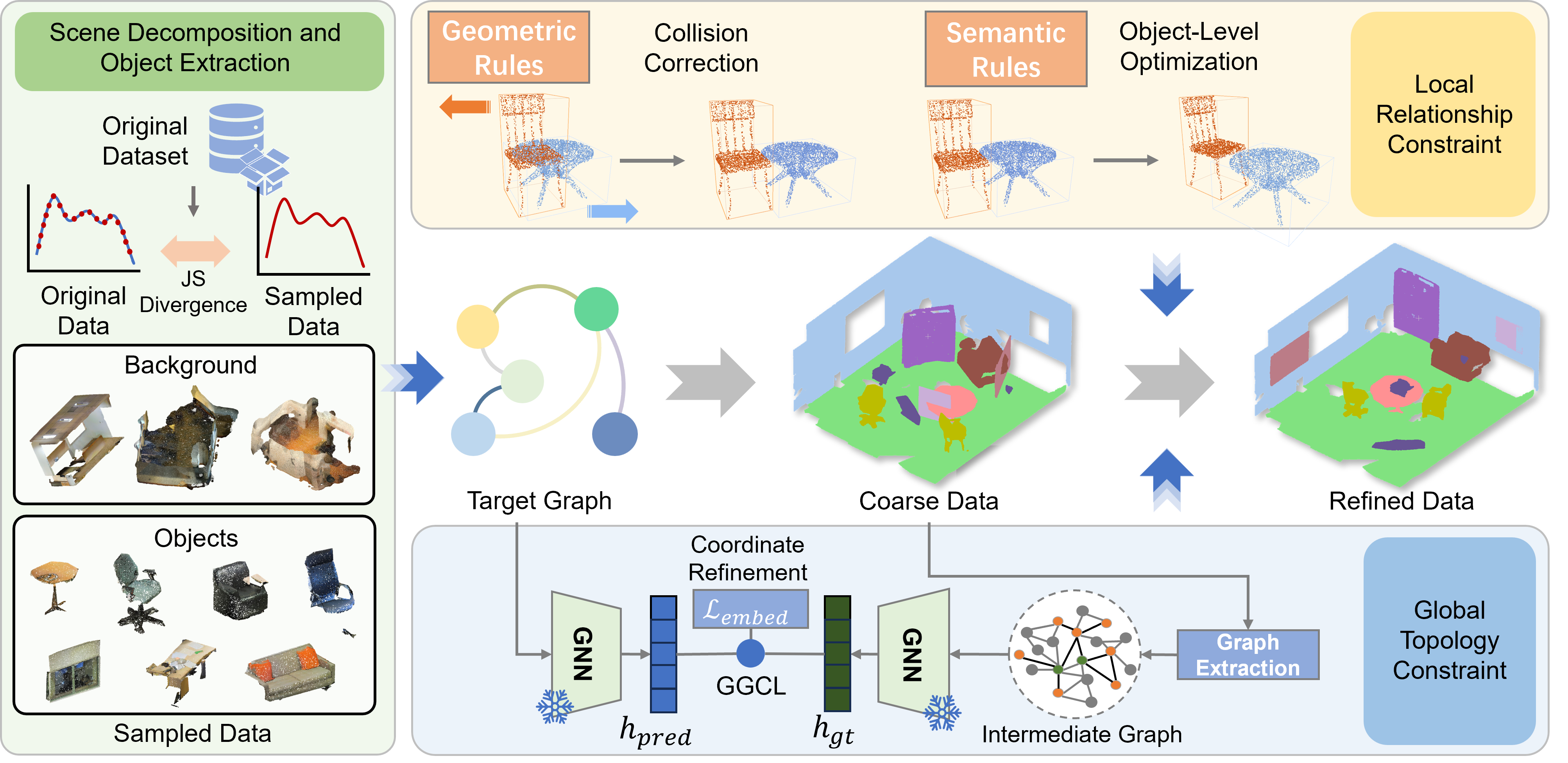}
    \vspace{-0.4cm}
    \caption{Overview of the proposed dual-level point cloud data augmentation framework. The pipeline consists of three key modules: (1) Scene Decomposition and Object Extraction, where the original dataset is decomposed into `Objects' and `Background' repositories. we perform sampling from the distribution—guided by JS divergence—to ensure the sampled distribution remains close to the original. The sampled elements are then used to generate a condition scene graph, serving as optimization guidance. (2) Local Relationship Constraints, which adjusts object positions and orientations according to geometric and semantic rules, including collision avoidance and relational constraints (e.g., "faces"). (3) Global Topology Constrains, where a pretrained Graph Neural Network (GNN) embeds both the ground-truth and predicted scene graphs to enforce structural consistency via a Graph Global Constraint Loss (GGCL) loss. This ensures the layout adheres to the intended relational structure. Together, these modules collaboratively generate diverse, semantically coherent, and geometrically valid point cloud scenes.}
    \label{fig:framework}
\vspace{-0.3cm}
\end{figure*}

Our framework synthesizes diverse and realistic 3D scenes by jointly enforcing local and global constraints. As shown in Fig.~\ref{fig:framework}, Section~\ref{subsec:decomposition} explains how scenes are decomposed into reusable background and foreground components. Section~\ref{subsec:graph_generation} details the construction of the Object Relationship Graph (ORG) to guide scene composition. Local-level geometric and semantic constraints are described in Sections~\ref{subsec:geometric_alignment} and~\ref{subsec:semantic_alignment}, while Section~\ref{subsec:topological_alignment} introduces global constraints via graph neural network embeddings. The complete generation pipeline is summarized in Section~\ref{subsec:generation_optimization}, where these constraints are jointly applied to ensure semantically coherent and structurally plausible 3D scenes.

\subsection{Scene Decomposition and Object Extraction}
\label{subsec:decomposition}

We propose a structured pipeline to decompose 3D scenes from both indoor datasets (ScanNet~\cite{dai2017scannet}, S3DIS~\cite{armeni20163d}) and outdoor datasets (Sem.KITTI~\cite{behley2019iccv}) into reusable semantic components. Given a raw 3D point cloud $\mathcal{P} \in \mathbb{R}^{N \times (3+C)}$, where $N$ denotes the number of points and $C$ represents additional features (e.g., RGB, normals), we leverage ground-truth segmentation labels to partition the scene into two parts: static background elements $\mathcal{B}$ (e.g., walls, floors, roads, buildings) and movable foreground objects $\mathcal{F}$ (e.g., furniture, vehicles, pedestrians). These decomposed components $\{P_k | k \in \mathcal{B} \}$ and $\{P_m | m \in \mathcal{F} \}$ provide a flexible repository for subsequent scene recomposition and augmentation. Please refer to the Appendix~\ref{append:dataset} for detailed statistical results.

\vspace{-0.3cm}
\subsection{Graph-Guided Scene Generation}
\label{subsec:graph_generation}
\begin{table}[t]
\centering
\resizebox{0.99\linewidth}{!}{%
  \renewcommand{\arraystretch}{2.5}
  \begin{tabular}{l | c}
    \toprule
    \textbf{Relation} & $\boldsymbol{P(\mathrm{relation}(A,B))}$ \\
    \midrule
    \textbf{Supported by}$(A,B)$ 
    & $\mathbf{1}\Bigl(\mathrm{overlap}_{xy}(A,B) > \tau\Bigr)
      \cdot
      \mathbf{1}\Bigl(\Delta z(A,B) \le \epsilon\Bigr)$ \\
      
    \textbf{Attached to}$(A,B)$ 
    & $\displaystyle \max\Biggl( \mathbf{1}\Bigl(\frac{|A \cap B|}{\min(|A|,|B|)} > \tau_{\text{att}}\Bigr),\; \mathbf{1}\Bigl(\mathcal{A}(\mathbf{d}_A, \mathbf{d}_B) > \tau_{\text{dir}}\Bigr) \Biggr)$ \\
    
    \textbf{Left of}$(A,B)$ 
    & $\mathbf{1}\Bigl(\frac{\mathrm{Vol}\bigl(A \cap \mathrm{left\_of}(B)\bigr)}{\mathrm{Vol}(A)} > \tau_{\text{left}}\Bigr)$ \\
    
    \textbf{Right of}$(A,B)$ 
    & $\mathbf{1}\Bigl(\frac{\mathrm{Vol}\bigl(A \cap \mathrm{right\_of}(B)\bigr)}{\mathrm{Vol}(A)} > \tau_{\text{right}}\Bigr)$ \\
    
    \textbf{Nearby}$(A,B)$ 
    & $\mathbf{1}\Bigl(\mathrm{dist}(A,B) \le t_{\mathrm{near}}\Bigr)$ \\
    
    \textbf{Faces}$(A,B)$ 
    & $\mathbf{1}\Bigl(\cos\!\Bigl(\mathrm{front}(A),\, \mathbf{c}(B) - \mathbf{c}(A)\Bigr) > \tau_{\text{face}}\Bigr)$ \\
    
    \textbf{Oriented with}$(A,B)$ 
    & $\mathbf{1}\Bigl(\mathrm{overlap}_{xy}(A,B) > \tau\Bigr)
      \cdot
      \mathbf{1}\Bigl(\cos(\mathbf{n}_A, \mathbf{n}_B) > \epsilon''\Bigr)$ \\
    \bottomrule
  \end{tabular}%
}
\caption{Formal definitions of spatial relationships between objects \(A\) and \(B\). Here, \(\mathbf{1}(\cdot)\) is the indicator function, \(\mathrm{overlap}_{xy}\) denotes the 2D horizontal overlap ratio, \(\Delta z\) represents the vertical distance between object bases, \(\mathbf{d}_A\) indicates the principal orientation vector, \(\mathbf{n}_A\) denotes the surface normal, \( |A \cap B| \) measures the 3D intersection volume, and \(\tau\), \(\epsilon\), \(\tau_{\text{att}}\), \(\tau_{\text{dir}}\), \(\tau_{\text{left}}\), \(\tau_{\text{right}}\), \(\tau_{\text{face}}\), and \(\epsilon''\) are tolerance thresholds.}
\label{tab:relations}
\vspace{-25pt}
\end{table}

To generate semantically consistent and diverse 3D environments, we construct an ruled-based \textit{Object Relationship Graph} (ORG), which models statistical co-occurrence patterns and spatial relationships observed in the source datasets. The complete pseudocode for ORG construction is provided in the Appendix \ref{pseudocode} for clarity. Given a set of extracted furniture instances $\mathcal{F}$ and background elements $\mathcal{B}$, the graph is defined as $\mathcal{G} = (\mathcal{V}, \mathcal{E}, W)$, where $\mathcal{V}$ denotes the set of object categories (including furniture and background elements), $\mathcal{E}$ denotes the set of edges capturing spatial relationships between object pairs, and $W$ encodes the corresponding connection strengths.

Each node $v_i \in \mathcal{V}$ represents an object class, and an edge $e_{ij} \in \mathcal{E}$ is established if a spatial relationship exists between objects $o_i$ and $o_j$ in the dataset. The adjacency matrix $A$ records the presence or absence of these relationships between object categories, where $A_{ij}=1$ indicates a valid relation between $o_i$ and $o_j$, and $A_{ij}=0$ otherwise. In our framework, the ORG is initialized with two primary structural nodes, \textit{floor} and \textit{wall}, serving as reference anchors for object placement.

To maintain consistency with the real-world data distribution, the activation probability of each object node during graph construction is regulated based on its occurrence frequency in the training dataset. Specifically, We employ category-wise Gaussian sampling to introduce instance-level randomness, while additionally incorporating a JS divergence\cite{lin1991divergence} regularization to globally align the generated node distribution with the empirical distribution observed in the source dataset.

Edge weights $w_{ij}$ quantify the co-occurrence strength between object categories $O_i$ and $O_j$, which are computed based on their frequency of simultaneous appearance in the source dataset $\mathcal{D}_{data}$. Specifically, $w_{ij}$ is defined as the normalized occurrence count of the object pair $(O_i, O_j)$ relative to all object pairs in the dataset:

\begin{equation}
    w_{ij} = \frac{\text{count}(O_i, O_j)}{\sum_{(O_m, O_n) \in \mathcal{D}_{data}} \text{count}(O_m, O_n)},
\end{equation}

\noindent where $\text{count}(O_i, O_j)$ denotes the number of times objects $O_i$ and $O_j$ appear together in the same scene. The resulting weight matrix $W$ is further normalized to $\tilde{W}$ for subsequent graph operations and sampling procedures.

\begin{equation}
    \tilde{W} = D^{-1/2} W D^{-1/2},
\end{equation}

where $D$ is the diagonal degree matrix with $D_{ii} = \sum_j W_{ij}$.

The relationships between objects are characterized based on empirical analysis of the source datasets. We define a set of canonical spatial relationships, such as those detailed in Table~\ref{tab:relations}. These relationships capture common interaction patterns like support, proximity, orientation, and relative positioning. During statistical analysis, if an object instance is found to have no defined relationship (\textit{none}) with any other object in a sampled scene context, it may be excluded from the co-occurrence statistics to avoid noise from potentially isolated or ambiguously placed objects.

In the augmentation phase, the ORG generation starts with the key context nodes. New object nodes and their relationships are sampled based on the learned co-occurrence probabilities, often modeled using probability distributions (e.g., derived from $w_{ij}$). For instance, we can sample new edges based on conditional probabilities $P(v_j | v_i)$:

\begin{equation}
    e_{ij} \sim \mathbb{1}(P(v_j | v_i) > \tau),
\end{equation}

\noindent where $\tau$ is a threshold controlling the density and diversity of generated graph structures. Additionally, to address potential imbalances or low segmentation accuracy for certain object classes, we can employ Ground-Truth sampling (GT Sampling), increasing the likelihood of including instances from underrepresented or challenging categories in the augmented scenes, thereby enhancing model robustness. Appendix~\ref{append:relationship_statistics} shows the statistics of relationships in each training datasets.

\subsection{Local Geometric Constraints}
\label{subsec:geometric_alignment}

Ensuring physically plausible object placement is critical for generating realistic 3D scenes that align with the geometric distribution observed in the real world. To enforce this structural coherence, we introduce constraints targeting fundamental geometric properties, primarily collision avoidance and surface alignment. These constraints guarantee that generated object configurations adhere to the distribution of physically valid arrangements, preventing interpenetration and ensuring stable orientations.

Collision avoidance is implemented using 3D bounding box intersection tests. Given two furniture objects $A_i$ and $A_j$ with bounding boxes $BB_i$ and $BB_j$, we define a collision penalty function as:

\begin{equation}
    L_{\text{collision}} = \sum_{i,j} \mathbb{1}(BB_i \cap BB_j \neq \emptyset) \cdot \text{vol}(BB_i \cap BB_j),
\end{equation}

\noindent where $\mathbb{1}(BB_i \cap BB_j \neq \emptyset)$ is an indicator function that activates when two bounding boxes intersect, and $\text{vol}(BB_i \cap BB_j)$ represents the overlapping volume. The objective is to minimize $L_{\text{collision}}$, reducing spatial conflicts.

Surface alignment ensures that objects are placed with appropriate orientations relative to surfaces. We perform plane detection using RANSAC and normal clustering, where each furniture object's principal axis is aligned using PCA. Given an object $o_i$ with normal $\mathbf{n}_i$ and an expected support surface normal $\mathbf{n}_s$, alignment is enforced by minimizing:

\begin{equation}
    L_{\text{alignment}} = \sum_{i} \left( 1 - \left| \mathbf{n}_i \cdot \mathbf{n}_s \right| \right),
\end{equation}

\noindent where the dot product $\mathbf{n}_i \cdot \mathbf{n}_s$ quantifies angular deviation. Objects such as tables and chairs are constrained to align with horizontal surfaces (e.g., floors), while smaller items like cups and pillows are positioned on top of furniture surfaces (e.g., tables or beds) using the same alignment mechanism.

By jointly optimizing these constraints during the layout refinement process, we steer the generated scenes toward the geometric distribution of physically plausible configurations, enhancing realism and consistency with real-world data.

\begin{figure}[t]
    \centering
    \includegraphics[width=0.95\linewidth]{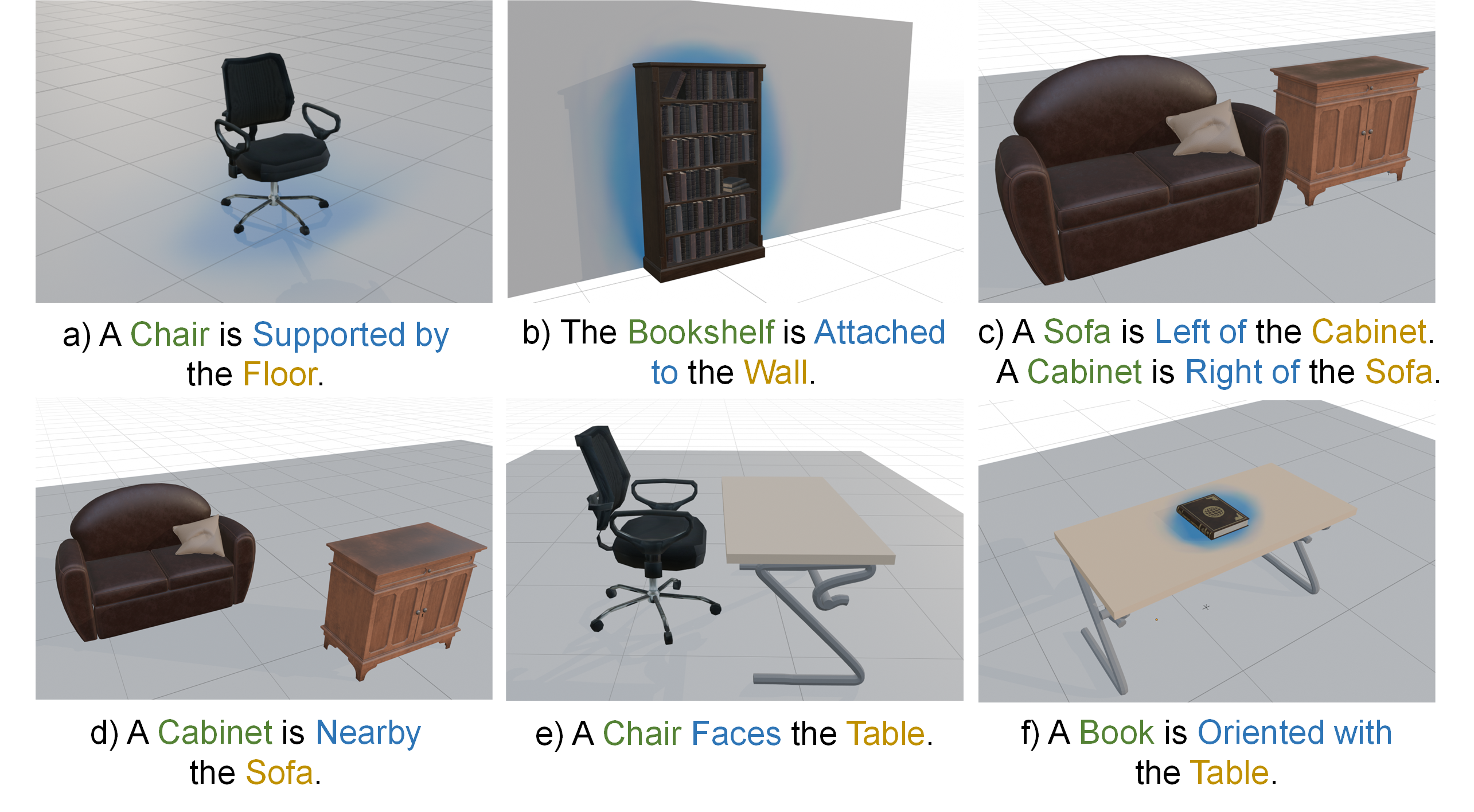}
    \vspace{-0.4cm}
    \caption{\textbf{Illustration of Semantic Rules for Object Placement in Scene Generation.} This diagram presents an array of semantic relationships.}
    \label{fig:rules}
\vspace{-0.5cm}
\end{figure}

\subsection{Local Semantic Constraints}
\label{subsec:semantic_alignment}

To establish functionally coherent arrangements among objects in the generated scenes, we define a set of canonical spatial relationships based on real-world interaction patterns. Each of the seven predefined spatial relationships (see Table~\ref{tab:relations}) is associated with a specific loss function that penalizes deviations from the anticipated spatial configurations. The overall semantic loss for a given scene is computed by summing the losses over all object pairs that have defined relational edges.

Let \( r(A,B) \) denote the relationship type between a pair of objects \( A \) and \( B \). The total semantic loss is expressed as:

\begin{equation}
L_{\text{semantic}} = \sum_{(A,B) \in \mathcal{R}} \alpha_{r(A,B)} L_{r(A,B)}(A,B),
\end{equation}

\noindent where \( \mathcal{R} \) represents the set of object pairs linked by a relationship, \( L_{r(A,B)} \) is the rule-specific loss function, and \( \alpha_{r(A,B)} \) is a weight controlling the influence of each relationship type.

The loss for the Supported By relation ensures horizontal overlap and minimal vertical offset:
\begin{align}
L_{\text{support}}(A,B) &= \lambda_{1}\, \max\!\Biggl(0,\; \frac{d_{\text{ce}}(A,B)}{\sqrt{\min\bigl(\text{Area}_{xy}(A),\text{Area}_{xy}(B)\bigr)}} - \tau \Biggr) \notag
\\
&+ \lambda_{2}\, \bigl|\Delta z(A,B) - \epsilon\bigr|.
\end{align}
Here, $\text{Area}_{xy}(A)$ denotes the 2D projected area of object $A$ onto the ground plane, and similarly for $B$. The function $d_{\text{ce}}(A,B)$ represents the centroidal Euclidean distance in the $xy$-plane between the projected footprints of objects $A$ and $B$.

For Attached To, the loss considers both intersection volume and orientation alignment:
\begin{equation}
L_{\text{attach}}(A,B) = \mu \left[1 - \max\left(\frac{|A \cap B|}{\min(|A|, |B|)}, \mathcal{A}(\mathbf{d}_A, \mathbf{d}_B)\right)\right]^2,
\end{equation}

\noindent Here, \(\mathcal{A}(\mathbf{d}_A, \mathbf{d}_B)\) is a continuous function that quantifies the alignment between the dominant direction vectors of \(A\) and \(B\), ensuring smooth gradient propagation during optimization. The directional relationships Left Of and Right Of penalize violations of spatial half-space alignment:
\begin{equation}
L_{\text{left}}(A,B) = \alpha_{1} \left[1 - \frac{\mathrm{Vol}(A \cap \mathrm{left\_of}(B))}{\mathrm{Vol}(A)}\right]^2,
\end{equation}
\begin{equation}
L_{\text{right}}(A,B) = \alpha_{2} \left[1 - \frac{\mathrm{Vol}(A \cap \mathrm{right\_of}(B))}{\mathrm{Vol}(A)}\right]^2.
\end{equation}

For proximity relations, the Nearby loss penalizes object pairs that are too distant from one another:
\begin{equation}
L_{\text{near}}(A,B) = \nu \max\left(0, \mathrm{dist}(A,B) - t_{\text{near}}\right).
\end{equation}

The Faces relation encourages objects to be oriented toward each other by maximizing directional alignment:
\begin{equation}
L_{\text{face}}(A,B) = \gamma \left[1 - \cos\left(\mathrm{front}(A), \mathbf{c}(B) - \mathbf{c}(A)\right)\right]^2.
\end{equation}

The Oriented With constraint jointly enforces horizontal overlap and parallel surface alignment:
\begin{align}
L_{\text{oriented}}(A,B) = \rho_{1} \max\left(0, \tau - \mathrm{overlap}_{xy}(A,B)\right) + \notag
\\
\rho_{2} \max\left(0, \epsilon'' - \cos(\mathbf{n}_A, \mathbf{n}_B)\right).
\end{align}

By minimizing the total semantic loss across all relationally connected object pairs, the optimization process enforces realistic, functional interactions and supports diverse yet coherent scene synthesis. These spatial constraints are integral to aligning generated content with high-level semantic distributions observed in real-world environments.

\subsection{Global Topological Constraints}
\label{subsec:topological_alignment}

While local geometric and semantic constraints effectively regulate pairwise relationships between objects, satisfying these local conditions alone does not guarantee that the overall scene layout adheres to the global structural patterns observed in real-world environments. In particular, scenes constructed purely based on object-wise constraints may still exhibit unreasonable global configurations, such as unrealistic clustering or sparse distributions of objects. To address this limitation, we introduce a global-level constraint that evaluates the holistic topological structure of the scene. Specifically, we compare the relational graph of the generated scene with the target Object Relationship Graph (ORG) sampled from real data (denoted as $\mathcal{G}_{\text{target}}$ in Sec.~\ref{subsec:graph_generation}), ensuring that the synthesized scene preserves both local interactions and global spatial organization consistent with real-world distributions.

We employ a pre-trained SceneGraphNet~\cite{zhou2019scenegraphnet} serving as a graph encoder for our global topological constraint. In our method, each node in the scene graph represents a furniture category (i.e., object label), and each edge corresponds to one of seven predefined spatial relationships between object pairs (such as \textit{supported by}, \textit{attached to}, \textit{left of}, \textit{right of}, etc.). Since the graph only encodes object categories and pairwise spatial relations, the required representational capacity is minimal and does not demand a complex architecture.

We define a \textbf{Graph Global Constraint Loss (GGCL)} that captures fine-grained discrepancies between the target and current graph structures. In our framework, the loss is formulated as follows:

\begin{align}
L_{\text{topology}} &= \lambda_{\text{ins}}\, N_{\text{ins}}(z_{\text{target}}, z_{\text{current}}) + \lambda_{\text{del}}\, N_{\text{del}}(z_{\text{target}}, z_{\text{current}}) \notag \\
&\quad + \lambda_{\text{sub}}\, \min_{\pi \in \Pi} \sum_{i \in \mathcal{M}} d_{\text{sub}} \Bigl( z_{\text{target}}^{(i)},\, z_{\text{current}}^{(\pi(i))} \Bigr) \notag \\
&\quad + \lambda_{\text{struct}}\, \|A_{\text{target}} - A_{\text{current}}\|_F,
\end{align}

\noindent Here, $N_{\text{ins}}$ and $N_{\text{del}}$ quantify the number of node insertions and deletions required to align the current graph embedding with the target, while $\min_{\pi \in \Pi} \sum_{i \in \mathcal{M}} d_{\text{sub}}( z_{\text{target}}^{(i)}, z_{\text{current}}^{(\pi(i))})$ computes the optimal substitution cost across all possible node matchings, with $d_{\text{sub}}(\cdot,\cdot)$ as the discrepancy function between node embeddings. The term $\|A_{\text{target}} - A_{\text{current}}\|_F$ measures the overall structural difference between the two graphs via the Frobenius norm of their adjacency matrices. The weighting coefficients $\lambda_{\text{ins}}$, $\lambda_{\text{del}}$, $\lambda_{\text{sub}}$, and $\lambda_{\text{struct}}$ balance these contributions. Importantly, the gradient of $L_{\text{topology}}$ is backpropagated not to update the fixed GNN weights, but to adjust the five degrees of freedom (DOF) pose parameters $(x_i, y_i, z_i, \theta_i, \phi_i)$ for each dynamic object $O_i$, thereby steering the scene toward the desired topological configuration. This mechanism explicitly steers the global layout towards the structural patterns characteristic of the target data distribution, as captured by the sampled ORG.

\subsection{Iterative Scene Generation and Optimization}
\label{subsec:generation_optimization}

The generation of each augmented 3D scene is progressively refined under the guidance of the dual-level constraints proposed in our framework. The process involves the following key steps:

First, a target Object Relationship Graph (ORG), denoted $\mathcal{G}_{\text{target}}$, is generated. The nodes representing object categories and the edges representing their relationships are stochastically activated based on co-occurrence statistics learned from the source dataset. This sampling process, potentially using metrics like Jensen-Shannon divergence to model similarity to the source distribution and assuming Gaussian properties for relationship likelihoods, produces a graph structure representative of plausible real-world scenes. Ground-Truth (GT) sampling strategies can be optionally integrated here to increase the frequency of specific object categories that may be underrepresented or challenging for downstream tasks.

Second, the scene is initialized. Dynamic object instances $P_m$ corresponding to the activated nodes in $\mathcal{G}_{\text{target}}$ are selected from the pool of extracted objects (Sec.~\ref{subsec:decomposition}) and placed into an initial, often random or heuristic, layout within the context defined by static background elements $P_k$.

Third, an iterative refinement process optimizes the poses of the dynamic objects. The optimization minimizes a total loss function $L_{\text{total}}$ that integrates the three levels of distribution alignment:

\begin{equation}
L_{\text{total}} = \lambda_{\text{geo}} L_{\text{geometric}} + \lambda_{\text{sem}} L_{\text{semantic}} + \lambda_{\text{topo}} L_{\text{topology}},
\end{equation}

\noindent where $L_{\text{geometric}}$ encompasses the collision and surface alignment losses (Sec.~\ref{subsec:geometric_alignment}), $L_{\text{semantic}}$ enforces pairwise object relationships based on the target ORG (Sec.~\ref{subsec:semantic_alignment}), and $L_{\text{topology}}$ aligns the global scene structure using GNN embeddings and GGCL (Sec.~\ref{subsec:topological_alignment}). The terms $\lambda_{\text{geo}}, \lambda_{\text{sem}}, \lambda_{\text{topo}}$ are hyperparameters balancing the contribution of each alignment level.

The optimization adjusts the 5-DOF pose $(x_i, y_i, z_i, \theta_i, \phi_i)$ for each dynamic object $O_i$ to find the configuration that minimizes $L_{\text{total}}$:

\begin{equation}
(x_i^*, y_i^*, z_i^*, \theta_i^*, \phi_i^*) = \arg\min_{(x_i, y_i, z_i, \theta_i, \phi_i)} L_{\text{total}}.
\end{equation}

The optimization process proceeds until a predefined convergence criterion is satisfied, such as the total loss falling below a threshold or the maximum number of iterations being reached. The resulting scene presents a novel yet reasonable configuration that inherits the structural characteristics and relational patterns observed in real-world environments.

%% file: sec/4_experiment.tex
\section{Experiment}
\label{sec:experiment}
\subsection{Experimental Setup}
\subsubsection{Datasets}
\label{subsec:datasets}

\begin{figure*}
    \centering
    \includegraphics[width=0.95\textwidth]{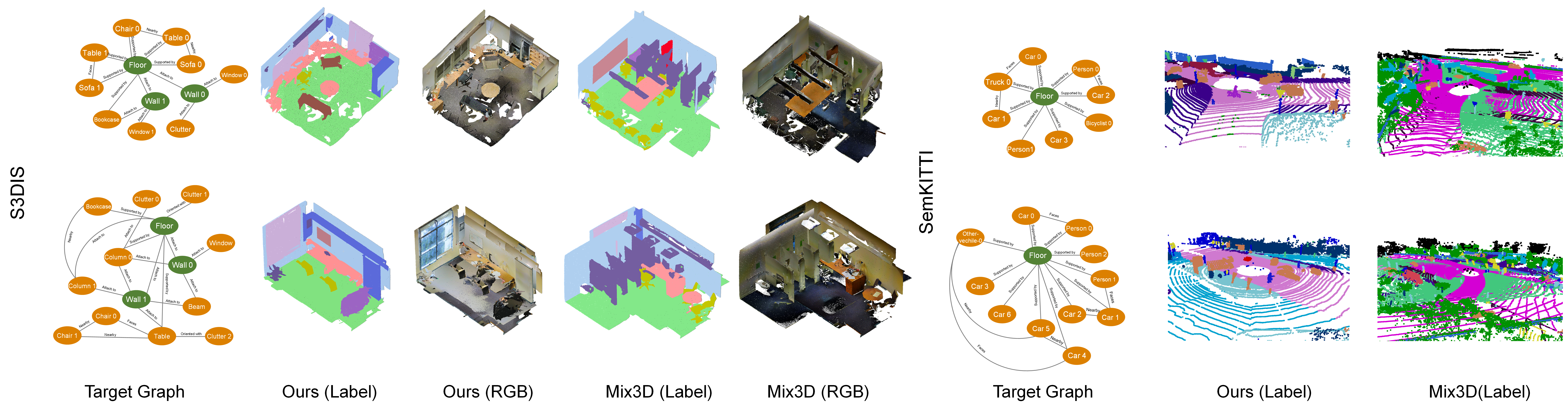}
    \vspace{-0.4cm}
    \caption{Visualization of augmented scenes generated by our method vs. Mix3D on S3DIS and SemanticKITTI datasets. The synthesized scenes are guided by the Object Relationship Graph (ORG) as the target graph, enabling the generation of novel layouts while preserving realistic spatial structures, while the layout generated by Mix3D is not reasonable}
    \label{fig:exp}
\vspace{-10pt}
\end{figure*}

\textbf{ScanNet}~\cite{dai2017scannet} is a widely used large-scale dataset containing 1,513 RGB-D reconstructed indoor scenes. It provides instance-level annotations for over 20 common indoor object categories. Following standard practice, we utilize 1,201 scenes for training and 312 scenes for validation and testing.
\textbf{S3DIS}~\cite{armeni20163d} offers detailed scans of indoor office environments across six large areas encompassing 272 rooms. Each point is annotated with XYZ coordinates, RGB color, and semantic labels covering 13 categories. We adopt a standard split, using Area 5 for testing and the remaining areas for training, resulting in 204 training rooms and 68 testing rooms.
\textbf{SemanticKITTI}~\cite{behley2019iccv} is a large-scale dataset providing dense point-wise semantic annotations for outdoor urban driving scenarios. We use the standard training and validation splits of SemanticKITTI~\cite{behley2019iccv}, covering 19 semantic classes commonly encountered in autonomous driving environments.

\vspace{-0.1cm}
\subsubsection{Data Preprocessing}
\label{subsec:preprocess}
For indoor datasets, we filter out incomplete, unlabeled, or isolated point clouds, as well as single-object point clouds lacking full room context. Occluded walls and floors, resulting from removing foreground objects such as furniture, are restored using Poisson surface reconstruction~\cite{kazhdan2006poisson} to maintain structural integrity (details in the Appendix~\ref{append:poisson} and results in the Appendix~\ref{append:poissonresult}). From these datasets, we extract a diverse set of boundary and furniture instances, totaling 505 and 203 boundary instances from ScanNet and S3DIS, respectively, along with 1,000 furniture instances from ScanNet and 2,000 from S3DIS.

In the context of the SemanticKITTI dataset, our preprocessing approach emphasizes the management of sequential LiDAR data through the analysis of individual scans or frames. We identify a total of 49,952 boundary instances and classify 179,092 static components, such as road surfaces and building facades, along with 195,382 dynamic elements, including vehicles and pedestrians, as foreground features. This thorough preprocessing yields a meticulously curated collection of reusable components, establishing a solid foundation for structured scene generation and augmentation in both indoor and outdoor settings.

\vspace{-0.1cm}
\subsubsection{Baseline}
To evaluate the effectiveness of our data augmentation approach, we integrate it into multiple state-of-the-art point cloud segmentation models and assess their performance with and without the inclusion of our generated data. We specifically focus on \textbf{OctFormer}~\cite{wang2023octformer} and \textbf{PTv3}~\cite{wu2024point}, both of which utilize transformer-based architectures renowned for their capability in large-scale 3D scene understanding. PTv3, in particular, is recognized for its superior performance, partly due to its integration of advanced data augmentation strategies such as Mix3D, CutMix, and PointAugment. These strategies make PTv3 an ideal baseline for evaluating the incremental benefits provided by our proposed method. By augmenting the training data with our generated scenes, we systematically evaluate segmentation accuracy and generalization improvements across both models.

\subsection{Result}
We evaluate the effectiveness of our data augmentation approach by measuring segmentation performance across different models and augmentation strategies on the ScanNet, S3DIS, and SemanticKITTI datasets. Table~\ref{tab:results} presents the quantitative comparison, where we report mean Intersection over Union (mIoU) scores for various model configurations. Appendix~\ref{append:gt_sampling} shows more details of GT Sampling.

\begin{table}[h]
    \centering
    \vspace{-0cm}
    % \setlength{\extrarowheight}{}
    % \resizebox{\linewidth}{!}{%
    \renewcommand{\arraystretch}{1}
    \begin{tabular}{l|c|c|c}
        \toprule
        \textbf{Method} & \textbf{ScanNet} & \textbf{S3DIS} & \textbf{Sem.KITTI} \\
        \midrule
        PointNeXt~\cite{qian2022pointnext} & 71.5 & 70.5 & - \\
        MinkUNet~\cite{choy20194d} & 72.2 & 65.5 & 63.8 \\
        SphereFormer~\cite{lai2023spherical} & - & - & 67.8 \\
        PTv2~\cite{wu2022point} & 75.4 & 71.6 & 70.3 \\
        OctFormer~\cite{wang2023octformer} & 74.6 & 67.1 & 60.3 \\
        OctFormer + Mix3D & 75.7 & 67.8 & 60.7 \\
        PTv3~\cite{wu2024point} & 78.6 & 74.7 & 72.3 \\
        \midrule
        \rowcolor{blue!10} OctFormer + Ours & \textbf{76.6} & \textbf{68.6} & \textbf{61.5} \\
        \midrule
        \rowcolor{blue!10} PTv3 + Ours & \textbf{79.8} & \textbf{75.5} & \textbf{73.2} \\
        \bottomrule
    \end{tabular}%
    % }
    \caption{Segmentation performance comparison (mIoU \%) on ScanNet, S3DIS, and KITTI.}
    \label{tab:results}
\vspace{-1cm}
\end{table}

To integrate our augmentation, we mix in additional synthesized data equivalent to 25\% of the original training set size for each dataset. This ensures that our generated scenes contribute meaningfully to model training while preserving the distributional characteristics of the original datasets. Fig.~\ref{fig:exp} presents a visualization of our augmentation data, demonstrating its diversity and structural coherence.

Our method consistently improves segmentation accuracy when integrated into segmentation models. For instance, \textbf{OctFormer+Ours} outperforms both \textbf{OctFormer} and \textbf{OctFormer+Mix3D}, demonstrating the advantage of our graph-guided augmentation in preserving scene structure. Similarly, our method further enhances performance across all three datasets for \textbf{PTv3}, which inherently incorporates Mix3D as part of its data augmentation strategy. These results validate that our approach effectively enhances the robustness of segmentation models by providing diverse yet semantically coherent training samples while maintaining spatial realism.

\vspace{-0.1cm}
\subsection{Ablation Study}
\label{subsec:ablation}
To evaluate the effectiveness of each component in our framework, we conduct an ablation study on the S3DIS and SemanticKITTI datasets, covering both indoor and outdoor scenarios. We compare our full model with several variants: (1) a naive baseline that randomly inserts additional data without any spatial constraints, (2) a variant with only geometric constraints, (3) a configuration applying all local-level constraints (including geometric and pairwise semantic relations) but without global structural guidance, and (4) the full model that integrates both local and global-level constraints. Additionally, we analyze the effect of augmentation scale by varying the ratio of generated data to 10\%, 25\%, and 50\% of the original dataset size.

\begin{table}[ht]
    \centering
    \vspace{-10pt}
    \setlength{\extrarowheight}{0pt}
    \resizebox{\linewidth}{!}{%
    \renewcommand{\arraystretch}{1}
    \begin{tabular}{p{5.5cm}|c|c}
        \toprule
        \textbf{Method} & \textbf{S3DIS} & \textbf{Sem.KITTI} \\
        \midrule
        Random Augmentation (25\%) & 57.2 & 59.9 \\
        Geometric Constraints Only (25\%) & 74.5 & 72.5 \\
        Local Constraints Only (25\%) & 75.1 & 72.9 \\
        Full Model (10\% Augmentation) & 75.1 & 72.8 \\
        Full Model (50\% Augmentation) & 73.9 & 71.6 \\
        \midrule
        \rowcolor{blue!10}Full Model (Local + Global Constraints, 25\%) &\textbf{75.5} & \textbf{73.2} \\
        \bottomrule
    \end{tabular}%
    }
    \caption{Ablation study results comparing different augmentation configurations and scales on S3DIS and SemanticKITTI.}
    \label{tab:ablation}
\vspace{-0.7cm}
\end{table}

As shown in Table~\ref{tab:ablation}, our full model integrating both local and global constraints achieves the best performance across S3DIS and SemanticKITTI. In contrast, randomly inserting data without constraints degrades performance due to spatially implausible scenes, while applying only geometric constraints leads to unstable results. Incorporating local-level semantic rules yields moderate improvements by enhancing pairwise relational coherence. The combination of local and global constraints achieves the highest accuracy, demonstrating the importance of hierarchical scene reasoning. Moreover, using 10\% augmented data already provides noticeable gains, while increasing the ratio to 50\% results in performance drop, likely due to distributional shift caused by excessive synthetic data. These findings suggest that moderate-scale, constraint-guided augmentation is most effective. Additionally, we evaluate GT Sampling, which increases the activation probability of the five worst-performing classes during generation. More details are provided in Appendix~\ref{append:gt_sampling}.

\subsection{t-SNE Visualization and Performance Analysis on Augmented Data}

To further evaluate the distributional properties of our generated data, we conduct t-SNE visualization and performance comparison experiments on the S3DIS dataset.

\begin{figure}
    \centering
    \includegraphics[width=0.80\linewidth]{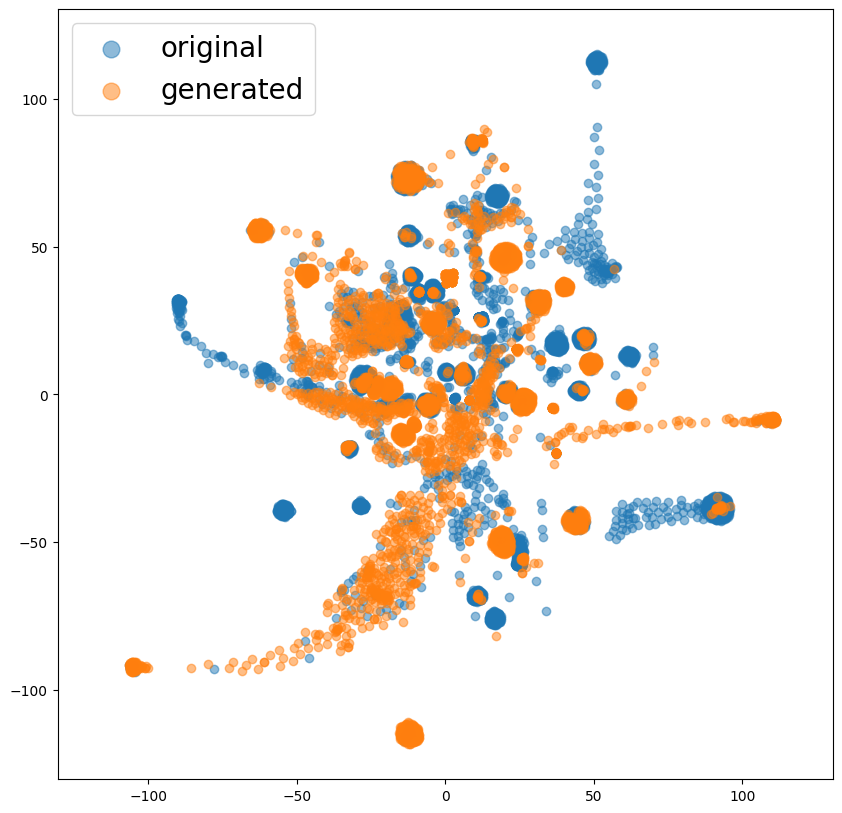}
    \vspace{-0.2cm}
    \caption{t-SNE visualization of features extracted from the last layer of PointNet++ on S3DIS, where blue denotes the original training data and orange represents our generated data. A higher color intensity indicates a greater density of samples in the corresponding region.}
    \label{fig:tsne}
    \vspace{-0.5cm}/
\end{figure}

For feature visualization, we extract high-level scene descriptors using a pre-trained PointNet++ model, utilizing the output from the final Set Abstraction (SA) layer. This layer aggregates contextual information from large spatial regions, implicitly encoding semantic content and global topological structures~\cite{qi2017pointnet++, ye2023closer}. The extracted features are projected into a 2D space using t-SNE~\cite{van2008visualizing}. As shown in Fig.~\ref{fig:tsne}, the feature distributions of our generated data closely align with those of the original training set, while also expanding into previously underrepresented regions. This confirms that our method enhances feature diversity without introducing distributional drift.

\begin{table}[h]
\centering
    %\setlength{\extrarowheight}{}
    %\resizebox{\linewidth}{!}{%
    \begin{tabular}{l|c|c|c|c|c|c}
        \toprule
        \multirow{2}{*}{Area} & \multicolumn{2}{c|}{Before Aug.} & \multicolumn{2}{c|}{After Aug.} & \multicolumn{2}{c}{Ours Only} \\
        \cline{2-7}
        & allAcc & mIoU & allAcc & mIoU & allAcc & mIoU \\
    \midrule
    Area 1 & 98.10 & 94.31 & 98.11 & 96.30 & 98.79 & 96.87 \\
    Area 2 & 98.19 & 93.02 & 98.17 & 96.42 & 98.33 & 96.44 \\
    Area 3 & 98.42 & 95.51 & 98.33 & 96.73 & 98.65 & 96.88 \\
    Area 4 & 98.26 & 94.04 & 98.28 & 96.64 & 98.81 & 97.01 \\
    Area 6 & 98.26 & 95.19 & 98.18 & 96.43 & 98.30 & 96.54 \\
    \rowcolor{blue!10}Area 5 & 92.45 & 74.68 & 93.05 & 75.51 & - & - \\
    \bottomrule
    \end{tabular}
    %}
    \caption{Comparison of segmentation performance (allAcc/mIoU \%) on each Area of S3DIS before and after our augmentation method. "Ours Only" denotes the evaluation results on generated data, not from training exclusively on generated data.}
\label{tab:s3dis_area}
\vspace{-0.7cm}
\end{table}

In addition, we report the segmentation performance of models trained on the original dataset and the augmented dataset (including the original dataset and our generated data), and evaluated on the generated data only across the six Areas in S3DIS. As shown in Table~\ref{tab:s3dis_area}, our method consistently improves allAcc and mIoU in Areas 1-4 and Area 6 after incorporating augmented data. For the unseen Area 5 (test set), performance is also improved, indicating enhanced generalization. Moreover, evaluating the generated data separately reflects the allAcc and mIoU performance on the augmented samples themselves. This comparison demonstrates the effectiveness of our augmentation method, highlighting improvements when training with augmented data versus the baseline.

%% file: sec/5_conclusion.tex
\section{Conclusion}
This work presents a graph-guided data augmentation framework that generates realistic and diverse 3D point cloud scenes through dual-level constraints. By explicitly modeling object co-occurrence statistics and enforcing both local-level geometric and semantic constraints and global-level topological consistency, our method enables the generation of high-quality synthetic scenes that better reflect real-world spatial patterns. Extensive experiments on both indoor and outdoor datasets demonstrate that our approach consistently improves segmentation performance across various models and datasets. Further analysis shows that our design, including GT sampling and global structure optimization, effectively enhances underrepresented categories and preserves meaningful scene layouts. In the future, we plan to extend our framework to more complex scene types and explore more efficient generation strategies to support large-scale applications, as well as real-time online generative augmentation methods.

%% file: sec/6_Acknowledgment.tex
\section{Acknowledgment}
This work is supported by the National Natural Science Foundation of China (No. 62406267).

%% file: sec/appendix.tex
\section{Appendix}
\label{sec:appendix}

\subsection{Dataset Processing Details}
\label{append:dataset}

To enable consistent scene decomposition across different datasets, we define a unified categorization strategy to partition each dataset into three sets: background elements for collision computation, floor elements for supporting objects, and movable foreground objects for scene recomposition.

For indoor datasets such as ScanNet~\cite{dai2017scannet} and S3DIS~\cite{armeni20163d}, we classify the \textit{wall} and \textit{floor} categories as static boundaries (\textbf{Background} and \textbf{Floor}), while all other categories are regarded as movable furniture instances (\textbf{Foreground}). This setting aligns with common indoor scene semantics where furniture placement dominates scene variability.

For the outdoor dataset SemanticKITTI~\cite{behley2019iccv}, we follow a similar principle but adjust the category assignments according to scene context. Specifically, we group \textit{road}, \textit{parking}, \textit{sidewalk}, \textit{other-ground}, and \textit{lane-marking} as \textbf{Floor} elements, which provide the supporting plane for dynamic objects. Categories such as \textit{building}, \textit{fence}, \textit{vegetation}, \textit{terrain}, and other static structures are grouped into the \textbf{Background} set, serving primarily as collision constraints. Movable object classes, including \textit{car}, \textit{bus}, \textit{person}, \textit{truck}, and their moving variants, are treated as \textbf{Foreground} instances subject to geometric and semantic optimization during scene generation.

\begin{table}[h]
\centering
\resizebox{\linewidth}{!}{
\begin{tabular}{l|c|c|c}
\hline
\textbf{Dataset} & \textbf{Floor} & \textbf{Background} & \textbf{Foreground} \\
\hline
ScanNet & 565 & 3078 & 7402 \\
S3DIS & 204 & 1203 & 5740 \\
SemanticKITTI & 19130 & 179092 & 195382 \\
\hline
\end{tabular}
}
\caption{Statistics of extracted elements for scene decomposition across datasets.}
\label{tab:data_statistics}
\end{table}

The number of extracted elements for each category in our decomposition process is summarized in Table~\ref{tab:data_statistics}.

It is important to note that the original implementation of OctFormer does not support S3DIS and SemanticKITTI. To enable a fair and consistent evaluation across datasets, we preprocess S3DIS and SemanticKITTI by converting them into a format compatible with the ScanNet data structure. This preprocessing step ensures that OctFormer can be trained and evaluated uniformly across all datasets considered in our experiments. In addition, we also incorporate Mix3D into OctFormer to assess its impact alongside our proposed method.

\subsection{Visualization of augmented data in ScanNet and STPLS3D}
\begin{figure*}
    \centering
    \includegraphics[width=0.8\textwidth]{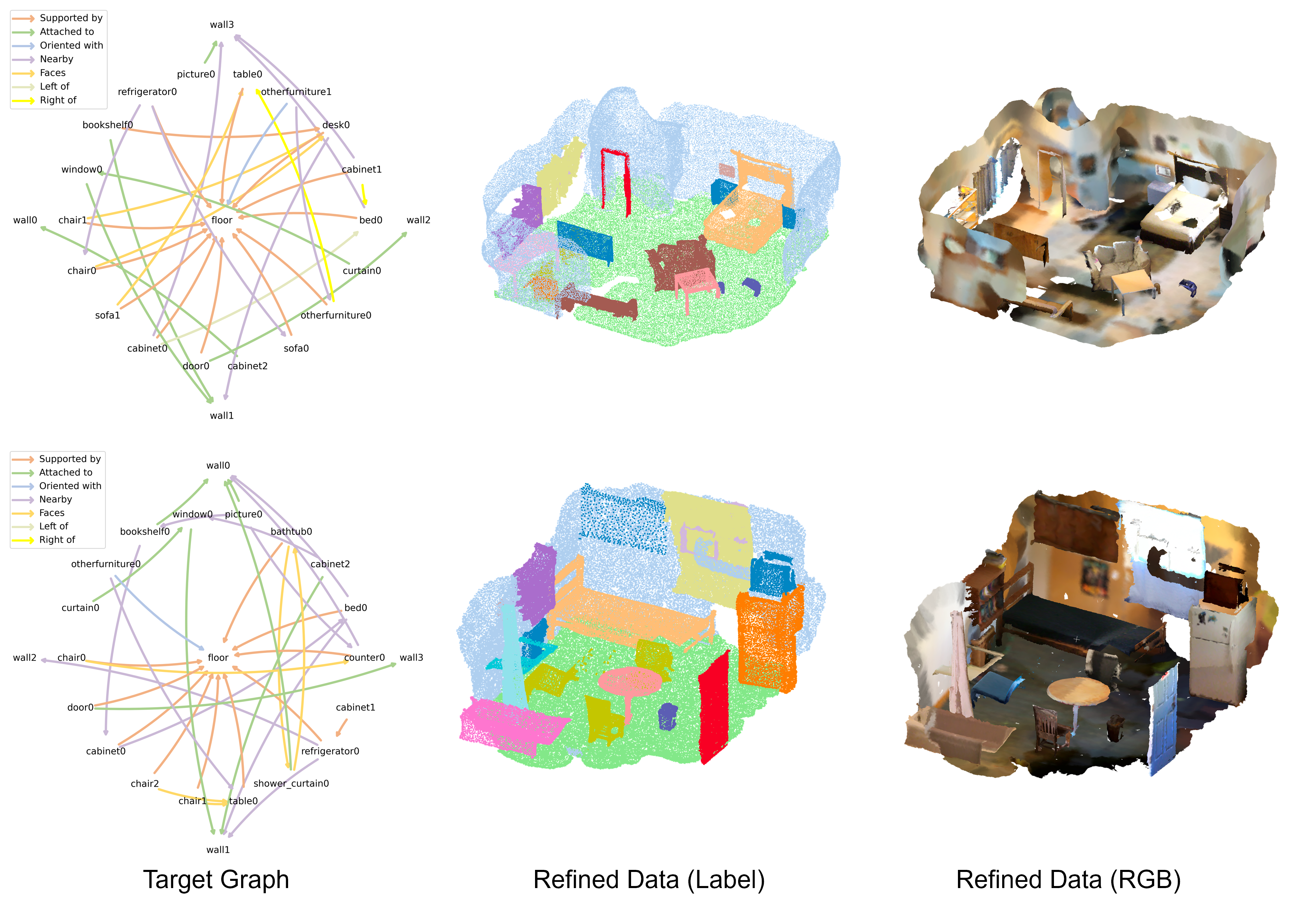}
    \caption{Visualization of augmented scenes generated by our method on ScanNet dataset. The synthesized scenes are guided by the Object Relationship Graph (ORG) as the target graph, enabling the generation of novel layouts while preserving realistic spatial structures.}
    \label{fig:scannet}
\end{figure*}

\begin{figure*}
    \centering
    \includegraphics[width=0.8\textwidth]{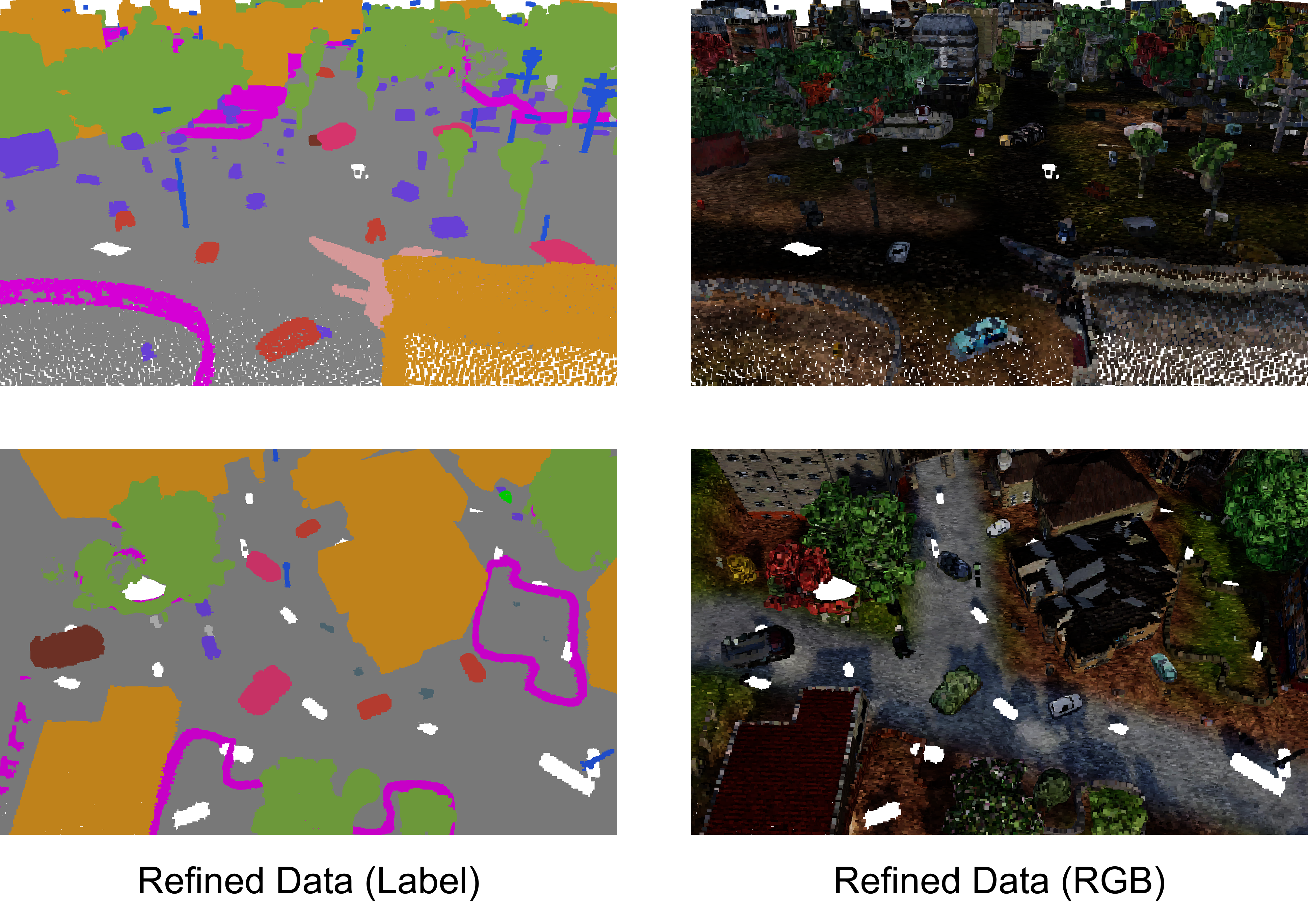}
    \caption{Visualization of augmented scenes generated by our method on STPLS3D dataset. As there are more than 300 nodes in each Object Relationship Graphs (ORG), the visualization of the graphs is impossible.}
    \label{fig:stpls3d}
\end{figure*}

Fig.~\ref{fig:scannet} and Fig.~\ref{fig:stpls3d} respectively show the visualization of augmented scenes generated by our method on ScanNet~\cite{dai2017scannet} and STPLS3D~\cite{chen2022stpls3d} datasets.

\begin{table}[ht]
    \centering
    \begin{tabular}{l|c|c|c}
        \toprule
        & wo Ours & Ours 10\% & Ours 25\% \\
        \midrule
        WMSC testing mIoU & 49.16 & 49.61 & 50.48 \\
        Synthetic V3 mIoU & 70.35 & 70.92 & 71.69 \\
        \bottomrule
    \end{tabular}
    \caption{Segmentation results on the STPLS3D dataset. ``WMSC testing mIoU'' refers to evaluation on the real-world WMSC test set after training on the synthetic subset, while ``Synthetic V3 mIoU'' refers to results on train/test splits within the Synthetic V3 subset. ``Ours 10\%'' and ``Ours 25\%'' denote experiments where 10\% and 25\% of the original data is augmented using our method, respectively.}
    \label{tab:stpls3d}
\end{table}

We further evaluate our method on the STPLS3D~\cite{chen2022stpls3d} dataset, as shown in Table~\ref{tab:stpls3d}. The results demonstrate that our augmentation framework consistently improves mIoU under both evaluation protocols, even in complex large-scale urban environments. On the WMSC test set, incorporating our augmented data yields improvements over the baseline, with mIoU rising from 49.16 to 50.48 when 25\% augmentation is applied. Similarly, on the Synthetic V3 split, our method raises mIoU from 70.35 to 71.69. These findings confirm the effectiveness and generalizability of our approach for 3D point cloud segmentation across challenging real-world scenes.

\subsection{Statistics of Spatial Relationships}
\label{append:relationship_statistics}

To construct semantically meaningful Object Relationship Graphs (ORGs) during scene generation, we conduct detailed statistical analysis of spatial relationships within the training sets of ScanNet~\cite{dai2017scannet}, S3DIS~\cite{armeni20163d}, and SemanticKITTI~\cite{behley2019iccv}. The statistics are used to guide both node sampling and edge relationship activation in the generated graphs.

\vspace{0.5em}
\noindent \textbf{Statistics Collection Protocol.}  
For each dataset, we first calculate the occurrence probability of each object category label within a scene by averaging its frequency across all training scenes. Subsequently, for each object instance, we identify its 10 nearest neighboring objects within the same scene based on Euclidean distance. We then compute the spatial relationship between the object and each of its neighbors, as well as between the object and static boundaries (floor, wall, or corresponding background classes in outdoor scenes). 

Importantly, spatial relationships such as \textit{left of} and \textit{right of} are not treated as symmetric. For example, if object $A$ considers object $B$ as one of its closest 10 neighbors and $B$ is located to the left of $A$, this relationship will be recorded as $left\_of(A,B)$. However, if $A$ does not appear within the closest 10 neighbors of $B$, the reverse relationship $right\_of(B,A)$ will not be recorded. This ensures the relationship statistics reflect realistic local observations rather than enforced symmetry.

\begin{figure}
    \centering
    \includegraphics[width=1\linewidth]{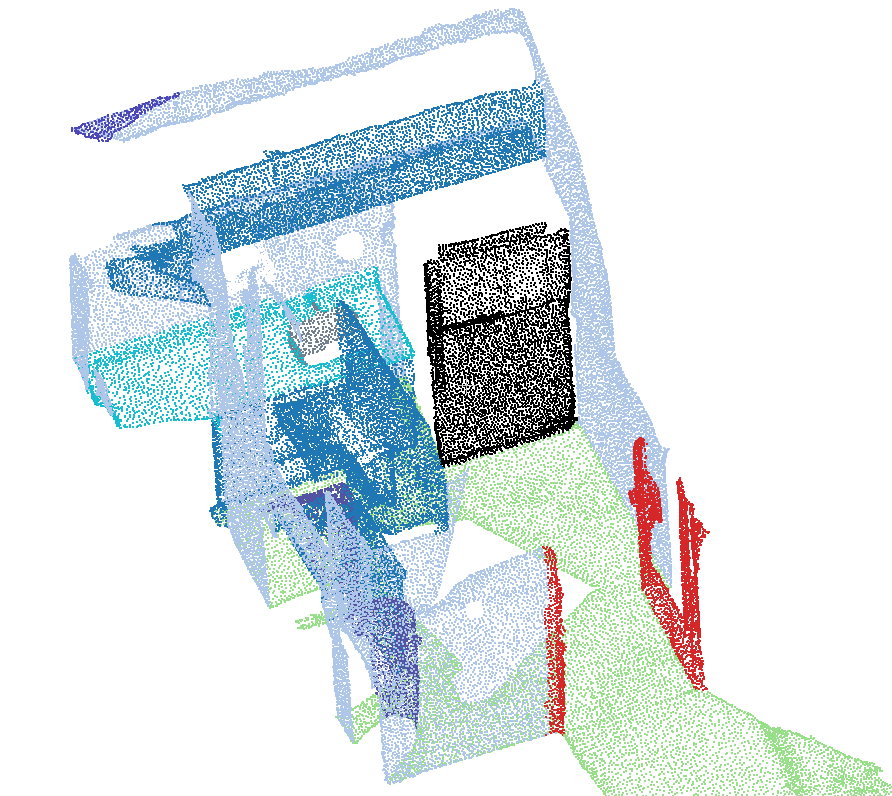}
    \caption{An example of an object with \textit{none} relationships to all surrounding furniture. The black point cloud in the figure represents a refrigerator. Due to incomplete scanning, only the front door of the refrigerator is captured. Such incomplete objects, which exhibit only a single surface rather than a complete 3D structure, often lead to difficulties in identifying spatial relationships. As a result, this object has a \textit{none} relationship with all surrounding furniture, walls, and floors.}
    \label{fig:none}
\end{figure}

\noindent \textbf{Graph Generation Strategy.}  
During scene generation, the ORG is constructed in two steps:

\textbf{1) Node Sampling.} Each object category's activation probability is modeled using a Gaussian distribution, where the mean is set to the average number of instances of that category observed per scene in the training set. For example, if chairs appear 3 times on average in a scene, their activation probability during graph generation follows a Gaussian distribution with a mean of 3. This allows a single object category to be activated multiple times within the same generated graph.

\textbf{2) Edge Activation.} Once nodes are sampled, edges between all node pairs are activated based on the empirical relationship probability distribution obtained from the training set. For instance, if the relationship between Chair and Table is observed to be \textit{faces} with 80\% probability and \textit{left of} with 10\% probability, we activate the edge type between Chair and Table in the generated graph according to a Gaussian function reflecting these probabilities. This edge sampling process ensures that different object categories have distinct and data-driven relational distributions.

\vspace{0.5em}
\noindent \textbf{Statistics of Relationships.}  
Table~\ref{tab:relationship_statistics} summarizes the number of instances and average occurrence per scene for each spatial relationship across the three datasets. In particular, we provide an example of a node with the \textit{None} relationship, as illustrated in Fig.~\ref{fig:none}.

\begin{table}[h]
\centering
\resizebox{\linewidth}{!}{
\begin{tabular}{l|c|c|c}
\hline
\textbf{Relationship} & \textbf{ScanNet} & \textbf{S3DIS} & \textbf{Sem.KITTI} \\
 & Total / Avg & Total / Avg & Total / Avg \\
\hline
Furniture Instances & 7402 / 13.10 & 5740 / 28.14 & 195382 / 10.21 \\
Supported By        & 5923 / 10.48 & 3856 / 18.90 & 185787 / 9.71  \\
Attached To         & 2238 / 3.96  & 1339 / 6.56  & 1169 / 0.06    \\
Left Of             & 866 / 1.53   & 659 / 3.23   & 27496 / 1.44   \\
Right Of            & 819 / 1.45   & 673 / 3.30   & 26200 / 1.37   \\
Nearby              & 1809 / 3.20  & 1141 / 5.59  & 41518 / 2.17   \\
Faces               & 1134 / 2.01  & 801 / 3.93   & 45199 / 2.36   \\
Oriented With       & 664 / 1.18   & 457 / 2.24   & 8806 / 0.46    \\
None                & 725 / 1.28   & 995 / 4.88   & 5288 / 0.28    \\
\hline
\end{tabular}
}
\caption{Statistics of spatial relationships across different datasets. We report both the total number of relationships and the average occurrence per scene.}
\label{tab:relationship_statistics}
\end{table}

This statistical analysis and the corresponding scene graph construction strategy ensure that our augmented scenes not only capture local geometric and semantic patterns but also respect dataset-specific global relational distributions, enabling the generation of diverse yet realistic 3D environments.

\subsection{Completion Method Based on Poisson Reconstruction}
\label{append:poisson}
The point cloud filling method commences with the crucial step of filtering the ground truth wall and floor data from the original point cloud dataset. This initial filtering process is of utmost significance as it lays the foundation for all subsequent operations, enabling a focused exploration of geometric features that are directly relevant to boundary construction and filling. By eliminating extraneous data, we streamline the analysis and ensure that our efforts are concentrated on the essential elements of the point cloud.

\begin{table*}[h]
    \centering
    \setlength{\extrarowheight}{0pt}
    \resizebox{\linewidth}{!}{%
    \renewcommand{\arraystretch}{1.2}
    \setlength{\tabcolsep}{4pt}
    \begin{tabular}{l|p{1.5cm}p{1.5cm}p{1.5cm}p{1.5cm}p{1.5cm}p{1.5cm}p{1.5cm}p{1.5cm}p{1.5cm}p{1.5cm}}
        \hline
        & \textbf{Wall} & \textbf{Floor} & \textbf{Cabinet} & \textbf{Bed} & \textbf{Chair} & \textbf{Sofa} & \textbf{Table} & \textbf{Door} & \textbf{Window} & \textbf{Bookshelf} \\
        \hline
        \textbf{Without Poisson} & 95.55 & 98.26 & 89.05 & 96.97 & 96.79 & 97.38 & 90.42 & 78.12 & 83.89 & 96.49 \\
        \textbf{With Poisson} & 95.28 & 98.30 & 88.33 & 96.28 & 97.08 & 97.02 & 90.21 & 81.59 & 85.26 & 96.56 \\
        \hline
        & \textbf{Picture} & \textbf{Counter} & \textbf{Desk} & \textbf{Curtain} & \textbf{Refrige-rator} & \textbf{Shower Curtain} & \textbf{Toilet} & \textbf{Sink} & \textbf{Bathtub} & \textbf{Other Furniture} \\
        \hline
        \textbf{Without Poisson} & 48.74 & 70.92 & 92.07 & 85.07 & 90.83 & 71.40 & 97.82 & 84.26 & 91.64 & 66.10 \\
        \textbf{With Poisson} & 47.16 & 82.72 & 91.93 & 88.37 & 92.07 & 81.72 & 97.97 & 81.35 & 92.65 & 71.73 \\
        \hline
        & \multicolumn{9}{c}{\textbf{Without Poisson}} & \textbf{79.49} \\
        & \multicolumn{9}{c}{\textbf{Overall mIoU}} & \textbf{79.79} \\
        \hline
    \end{tabular}%
    }
    \caption{Segmentation performance (mIoU \%) comparison on ScanNet with and without Poisson surface reconstruction.}
    \label{tab:poisson}
\end{table*}

Subsequently, leveraging the filtered ground truth (GT) floor and wall points as a reliable reference, we embark on a search for additional points within the scene to construct a preliminary, or coarse, boundary. Given the common occurrence of occlusion in real-world scenarios, the GT boundary is often incomplete. To address this challenge, we introduce the innovative concept of the \textit{fake boundary}. To generate this, we first construct a KD-tree for the raw data. A KD-tree, a sophisticated space-partitioning data structure, offers remarkable efficiency in performing nearest neighbor searches within the three-dimensional $(x, y, z)$ point cloud space. This data structure significantly accelerates the search process, making it possible to handle large-scale point cloud datasets in a computationally feasible manner.
We then utilize the GT boundary as a query to identify points that satisfy two specific conditions: 1) The Euclidean distance condition: the Euclidean distance $d$ from a point $p=(x_p, y_p, z_p)$ in the raw data to the GT boundary must be less than $\mu$, expressed mathematically as $d(p, boundary_{GT}) < \mu$. Here, if $q = (x_q, y_q, z_q)$ is a point on the GT boundary, the Euclidean distance: 
\begin{equation}
d = \sqrt{(x_p - x_q)^2+(x_p - x_q)^2 + (x_p - x_q)^2}
\end{equation}
2) The normal vector angle condition: The angular difference $\alpha$ between the normal vector $\vec{n}_p$ of point $p$ and the normal vector $\vec{n}_{GT}$ of the GT boundary should less than $\theta$. This angular difference is calculated using the dot product formula:
\begin{equation}
\cos(\alpha) = \frac{\vec{n}_p\cdot \vec{n}_{GT}}{|\vec{n}_p|\cdot |\vec{n}_{GT}|}
\end{equation}
and we enforce the constraint $ \alpha < \theta$.

Upon completion of this search, the retrieved points form a coarse boundary, which serves as the input for the subsequent Poisson Surface Reconstruction. Poisson Surface Reconstruction, a powerful technique for filling holes in the coarse boundary, is grounded in the solution of a Poisson equation. Given a set of points with associated normal vectors, the objective is to determine a smooth surface $S$ that passes through these points. Mathematically, considering a signed distance function $f(x)$ with $x\in \mathbb{R}^3$ , the surface $S$ is defined as the zero-level set of $f(x)$. The Poisson equation for surface reconstruction is $\Delta f = \rho$, where $\Delta$ represents the Laplace operator:
\begin{equation}
\Delta = \frac{\partial^2}{\partial x^2} + \frac{\partial^2}{\partial y^2} + \frac{\partial^2}{\partial z^2}
\end{equation}
and $\rho$ is a source term intricately related to the input points and their normals.

\begin{figure}[t]
    \centering
    \includegraphics[width=\linewidth]{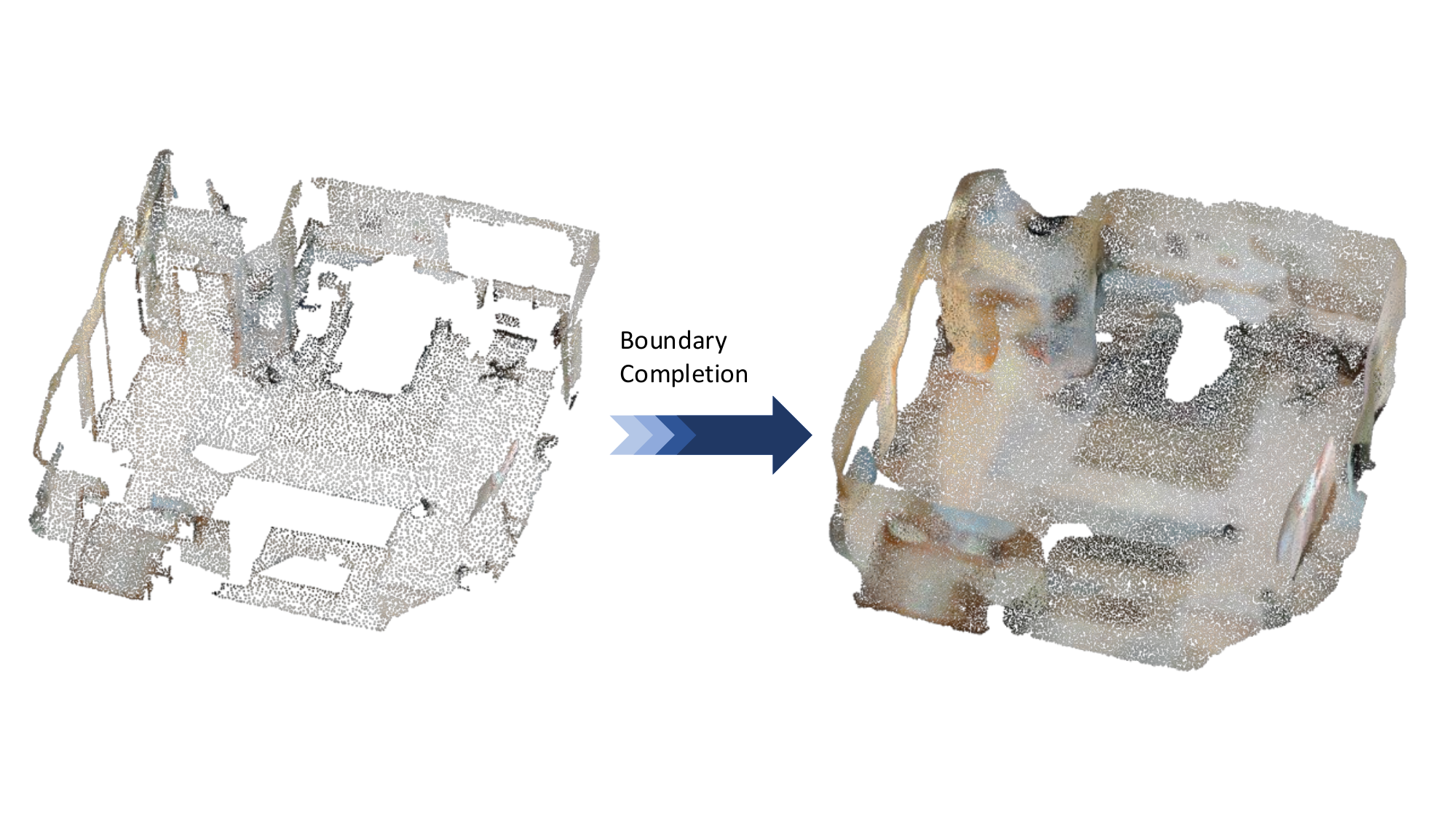}
    \caption{Complete the boundary in Scannet}
    \label{fig:completion}
\end{figure}

\begin{table*}[h]
    \centering
    \setlength{\extrarowheight}{0pt}
    \resizebox{\linewidth}{!}{%
    \renewcommand{\arraystretch}{1.2}
    \setlength{\tabcolsep}{4pt}
    \begin{tabular}{l|p{1.5cm}p{1.5cm}p{1.5cm}p{1.5cm}p{1.5cm}p{1.5cm}p{1.5cm}p{1.5cm}p{1.5cm}p{1.5cm}}
        \hline
        & \textbf{Wall} & \textbf{Floor} & \textbf{Cabinet} & \textbf{Bed} & \textbf{Chair} & \textbf{Sofa} & \textbf{Table} & \textbf{Door} & \textbf{Window} & \textbf{Bookshelf} \\
        \hline
        \textbf{Before GT-S} & 95.08 & 98.40 & 84.37 & 95.94 & 96.18 & 97.65 & 89.82 & 82.01 & 85.77 & 96.54 \\
        \textbf{After GT-S} & 95.28 & 98.30 & 88.33 & 96.28 & 97.08 & 97.02 & 90.21 & 81.59 & 85.26 & 96.56 \\
        \hline
        & \cellcolor{blue!10}\textbf{Picture} & \textbf{Counter} & \textbf{Desk} & \textbf{Curtain} & \cellcolor{blue!10}\textbf{Refrige-rator} & \cellcolor{blue!10}\textbf{Shower Curtain} & \textbf{Toilet} & \cellcolor{blue!10}\textbf{Sink} & \textbf{Bathtub} & \cellcolor{blue!10}\textbf{Other Furniture} \\
        \hline
        \textbf{Before GT-S} & \cellcolor{blue!10}44.09 & 80.85 & 86.78 & 84.47 & \cellcolor{blue!10}66.24 & \cellcolor{blue!10}82.07 & 98.11 & \cellcolor{blue!10}79.80 & 92.84 & \cellcolor{blue!10}68.40 \\
        \textbf{After GT-S} & \cellcolor{blue!10}47.16 & 82.72 & 91.93 & 88.37 & \cellcolor{blue!10}82.07 & \cellcolor{blue!10}81.72 & 97.97 & \cellcolor{blue!10}84.35 & 92.65 & \cellcolor{blue!10}74.73 \\
        \hline
        & \multicolumn{9}{c}{\textbf{Before GT-S}} & \textbf{79.55} \\
        & \multicolumn{9}{c}{\textbf{Overall mIoU}} & \textbf{79.79} \\
        \hline
    \end{tabular}%
    }
    \caption{Segmentation performance (mIoU \%) comparison on ScanNet with and without GT Sampling. Categories selected for GT Sampling show clear performance improvement. The highlighted categories indicate those augmented with GT Sampling.}
    \label{tab:gtsampling}
\end{table*}

In practical implementation, the operation steps are as follows. First, a volumetric grid enclosing the point cloud data is defined. This grid serves to discretize the 3D space, dividing it into a series of smaller, manageable cells. For each point $p$ in the coarse boundary, a value $v$ is assigned to the corresponding grid cells. This value is determined based on a combination of the distance $d$ from the point to the grid cell center $c = (x_c,y_c, z_c)$ and the orientation of the normal vector $\vec{n}_p$.  A commonly employed approach is $v = \frac{\vec{n}_p\cdot \vec{r}}{d}$, where $\vec{r} = (x_p-x_c, y_p-y_c, z_p - z_c)$.

Subsequently, the discrete Poisson equation is solved on the grid using numerical methods such as the conjugate gradient method. This iterative process adjusts the values of the grid cells in a systematic manner to find the function $f(x)$ that best satisfies the Poisson equation under the given boundary conditions. After Poisson Surface Reconstruction, although the obtained boundary is complete, it often exhibits a regular point distribution that differs from real-world data. To rectify this, we introduce perturbations. For a point $p = (x_p, y_p, z_p)$ on the boundary, the perturbed point:
\begin{equation}
p' = (x_p + \epsilon_x,y_p + \epsilon_y,z_p + \epsilon_z )
\end{equation}
is generated, where $\epsilon$ is random values drawn from a Gaussian distribution $N(0, \sigma^2)$ with a mean of 0 and a small standard deviation $\sigma$. The resulting filled boundary can then be utilized as a fundamental building block for applications such as point cloud generation or 3D model reconstruction.

\subsection{Effect of Poisson Surface Reconstruction}
\label{append:poissonresult}

\begin{figure}
    \centering
    \includegraphics[width=1\linewidth]{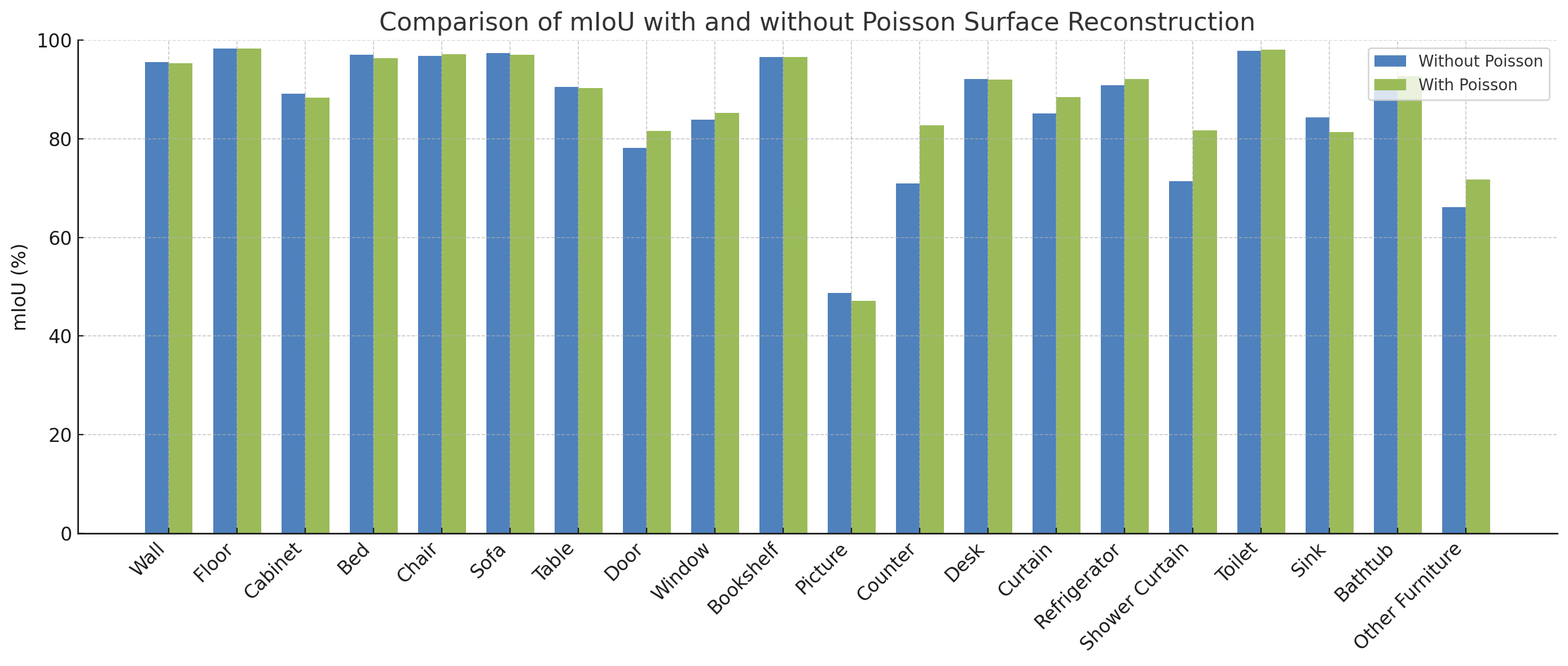}
    \caption{Comparison of mloU(\%) with and without Poisson Surface Reconstruction}
    \label{fig:poisson}
\end{figure}

To evaluate the impact of Poisson surface reconstruction in our data augmentation pipeline, we compare segmentation performance on PTV3 trained with augmented data both with and without hole-filling. Specifically, we analyze the effect of restoring occluded floors and walls after furniture removal. Table~\ref{tab:poisson} and Fig.~\ref{fig:poisson} presents the per-category and overall mean Intersection over Union (mIoU) scores on the ScanNet dataset.

The results indicate that Poisson reconstruction leads to a slight overall improvement in segmentation accuracy, with mIoU increasing from 78.19\% to 78.79\%. While certain categories, such as \textbf{counter} and \textbf{shower curtain}, show significant gains, others, including \textbf{cabinet} and \textbf{picture}, experience minor decreases. Notably, the categories that exhibit the most improvement—such as \textbf{counter, shower curtain}, and \textbf{other furniture}—are those that frequently interact with walls or floors. This suggests that Poisson surface reconstruction enhances segmentation performance particularly for objects that rely on well-defined boundary conditions. However, in categories where the original occlusions were minimal, the reconstruction may introduce slight inconsistencies. These findings highlight the trade-off between geometric consistency and segmentation accuracy, demonstrating that Poisson surface reconstruction generally enhances model robustness in indoor scene understanding, particularly in boundary-sensitive regions.

To further improve the performance of categories that are difficult to segment, we incorporate a Ground-Truth (GT) Sampling strategy during scene generation. This strategy aims to mitigate the long-tail problem commonly observed in indoor point cloud segmentation, where certain object categories appear infrequently or exhibit lower segmentation accuracy.

Specifically, we first analyze the validation results of the baseline segmentation model (PTv3) on the ScanNet dataset. We identify the five worst-performing categories in terms of mean Intersection-over-Union (mIoU): \textit{picture}, \textit{refridgerator}, \textit{otherfurniture}, \textit{sink}, and \textit{counter}. During ORG generation, the activation probability of these categories' nodes is increased to three times their original values. This encourages the generated scenes to include more instances of these challenging categories, thereby providing richer supervision for the segmentation model.

\subsection{Effectiveness of GT Sampling}
\label{append:gt_sampling}

\begin{figure}
    \centering
    \includegraphics[width=1\linewidth]{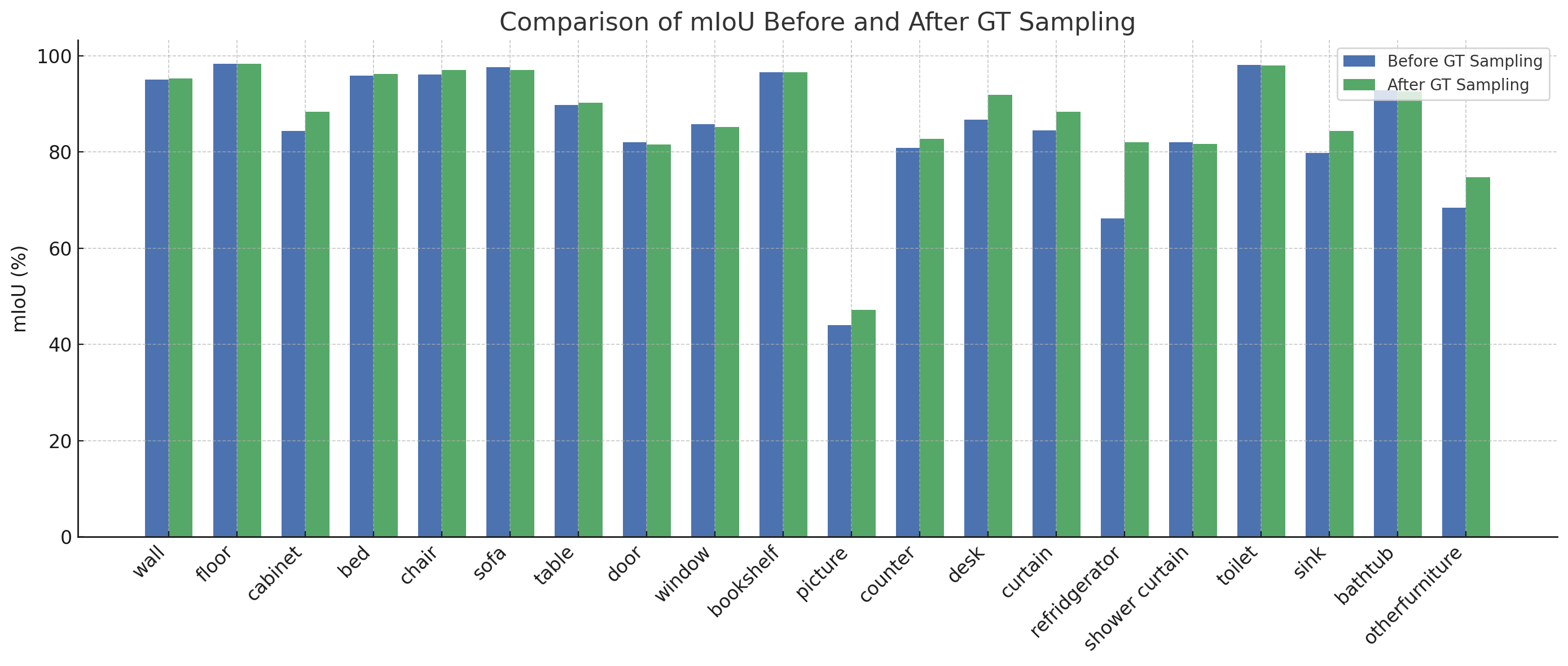}
    \caption{Comparison of mloU(\%) Before and After GT Sampling}
    \label{fig:gt}
\end{figure}

Table~\ref{tab:gtsampling} and Fig.~\ref{fig:gt} reports the segmentation results before and after applying GT sampling. We observe that the mIoU of the difficult categories improves significantly after applying this strategy. For example, \textit{picture} improves from 44.09\% to 47.16\%, \textit{refridgerator} improves from 66.24\% to 82.07\%, and \textit{otherfurniture} improves from 68.40\% to 74.73\%. Furthermore, we find that the GT-sampling (GT-S) strategy has negligible impact on the performance of already well-performing categories, indicating that our method mainly enhances the representation of rare or hard-to-segment classes without introducing noise to the overall scene distribution. Overall, the mIoU improves from 79.55\% to 79.79\%, demonstrating the effectiveness of our GT sampling design.

\subsection{Pseudocode of Object Relationship Graph Generation}
\label{pseudocode}

\begin{algorithm}[hb]
\caption{Object Relationship Graph (ORG) Generation}
\label{alg:org_generation}
\begin{algorithmic}[1]
\REQUIRE Training dataset $\mathcal{D}_{\text{data}}$, categories $\mathcal{C}$, spatial relationship rules $\mathcal{R}$, node activation means $\mu_c$ for $c \in \mathcal{C}$, JS divergence regularization, edge co-occurrence statistics
\ENSURE Object Relationship Graph $\mathcal{G} = (\mathcal{V}, \mathcal{E}, W)$
\STATE Initialize nodes $\mathcal{V} \gets \{\text{floor}, \text{wall}\}$
\FOR{each category $c \in \mathcal{C}$}
    \STATE Sample number of instances $n_c$ for $c$ using Gaussian with mean $\mu_c$
    \STATE Apply JS divergence regularization to align node counts with dataset distribution
    \STATE Add $n_c$ nodes of category $c$ to $\mathcal{V}$
\ENDFOR
\FOR{each node pair $(v_i, v_j)$ in $\mathcal{V}$}
    \STATE Compute $p_{ij}$ as the empirical co-occurrence probability of $(c_i, c_j)$ estimated from $\mathcal{D}_{\text{data}}$
    \STATE Sample spatial relationship $r_{ij}$ from rules $\mathcal{R}$ based on $p_{ij}$
    \IF{$r_{ij} \neq \text{none}$ and $p_{ij}$ exceeds threshold}
        \STATE Add edge $e_{ij}$ of type $r_{ij}$ to $\mathcal{E}$
        \STATE Set edge weight $w_{ij}$ as $p_{ij}$
    \ENDIF
\ENDFOR
\STATE Construct weighted adjacency matrix $W$ and normalize to $\tilde{W}$
\RETURN $\mathcal{G} = (\mathcal{V}, \mathcal{E}, W)$
\end{algorithmic}
\end{algorithm}